\documentclass{article}

\usepackage[preprint]{neurips_2022}

\usepackage{paralist}
\usepackage{algorithm}
\usepackage{algpseudocode}
\usepackage{graphicx}
\usepackage{natbib}
\usepackage{wrapfig}
\usepackage{floatflt}
\usepackage{graphics}

\usepackage[nowatermark]{fixmetodonotes}

\usepackage{amsmath,amsfonts,bm}

\def\eqref#1{equation~\ref{#1}}

\def\1{\bm{1}}

\def\rvw{{\mathbf{w}}}

\def\vzero{{\bm{0}}}

\def\vb{{\bm{b}}}

\def\ve{{\bm{e}}}

\def\vw{{\bm{w}}}
\def\vx{{\bm{x}}}

\def\evw{{w}}

\DeclareMathAlphabet{\mathsfit}{\encodingdefault}{\sfdefault}{m}{sl}
\SetMathAlphabet{\mathsfit}{bold}{\encodingdefault}{\sfdefault}{bx}{n}

\def\sR{{\mathbb{R}}}

\usepackage{hyperref}
\hypersetup{colorlinks,linkcolor={blue},citecolor={blue},urlcolor={blue}}
\usepackage{url}

\algnewcommand\AND{\textbf{and} }

\usepackage{caption}
\usepackage{calc}
\usepackage{cutwin}

\newcommand\picwin[3][0]{ 
\newsavebox\wpstuff \savebox{\wpstuff}{\parbox{#2}{\centering #3}}
\opencutright
\def\windowpagestuff{\flushright\usebox{\wpstuff}}
\newlength\hhh \settototalheight{\hhh}{\usebox{\wpstuff}}
\newlength\www \setlength{\www}{\dimexpr\textwidth-#2-1em\relax}
\begin{cutout}{#1}{\www}{0pt}{\the\numexpr\hhh/\baselineskip+1\relax}
\end{cutout}
}

\usepackage[utf8]{inputenc} 
\usepackage[T1]{fontenc}    
\usepackage{hyperref}       
\usepackage{url}            
\usepackage{booktabs}       
\usepackage{multirow}
\usepackage{amsfonts}       
\usepackage{nicefrac}       
\usepackage{microtype}      
\usepackage{xcolor}         
\usepackage{makecell}

\newtheorem{myProp}{Property}

\graphicspath{
{Figures/}
}

\title{Geometrically Guided Integrated Gradients}

\author{Md Mahfuzur Rahman, Noah Lewis \& Sergey Plis \thanks{ https://trendscenter.org/} \\
Tri-institutional Center for Translational Research in Neuroimaging and Data Science (TReNDS)\\
Georgia State University, Georgia Institute of Technology, Emory University\\
Atlanta, GA, USA \\
\texttt{\{mahfuz.gsu, lhd231, s.m.plis\}@gmail.com}
}

\begin{document}

\maketitle

\begin{abstract}
Interpretability methods for deep neural networks mainly focus on the sensitivity of the class score with respect to the original or perturbed input, usually measured using actual or modified gradients. Some methods also use a model-agnostic approach to understanding the rationale behind every prediction. In this paper, we argue and demonstrate that local geometry of the model parameter space relative to the input can also be beneficial for improved post-hoc explanations. To achieve this goal, we introduce an interpretability method called ‘geometrically-guided integrated gradients' that builds on top of the gradient calculation along a linear path as traditionally used in integrated gradient methods. However, instead of integrating gradient information, our method explores the model's dynamic behavior from multiple scaled versions of the input and captures the best possible attribution for each input. We demonstrate through extensive experiments that the proposed approach outperforms vanilla and integrated gradients in subjective and quantitative assessment. We also propose a "model perturbation" sanity check to complement the traditionally used "model randomization" test.
\end{abstract}

\section{Introduction}

The past decade has seen a wealth of new advancements in deep learning (DL), improving performance in a wide array of possible problems in many areas, especially in classification tasks from computer vision and natural language processing~\citep{chai2021deep, young2018recent}. However, this improved performance comes with a cost, the models are not easily explained or described with simple observations. This lack of intelligibility hinders wide-scale use of these models in safety-critical domains such as healthcare, education, the justice system, and many others. This need for decipherability has given way to parallel advancements in a subfield of machine learning known as interpretability. Interpretability has grown as quickly as the DL field itself, providing many new state-of-the-art techniques \citep{baehrens2010explain, simonyan2013deep,
    shrikumar2017learning,
    sundararajan2017axiomatic, selvaraju2017grad,
     bach2015pixel, montavon2017explaining, kapishnikov2019xrai, kapishnikov2021guided}

Interpretability research generally tackles these problems by either describing the model’s decision-making processes or by generating post-hoc explanations, either for the model as a whole, or for each sample. There are many post-hoc explanation methods from research spanning the last 2 decades. The performance and properties of these methods vary widely across different architectures and domains. In recent years, beginning with integrated gradients (IG)~\citep{sundararajan2017axiomatic}, some of these methods \citep{kapishnikov2019xrai, kapishnikov2021guided, xu2020attribution} have satisfied two important properties: \emph{sensitivity} and \emph{implementation invariance}.

\begin{figure}[!ht]
    \includegraphics[width=1.0\linewidth]{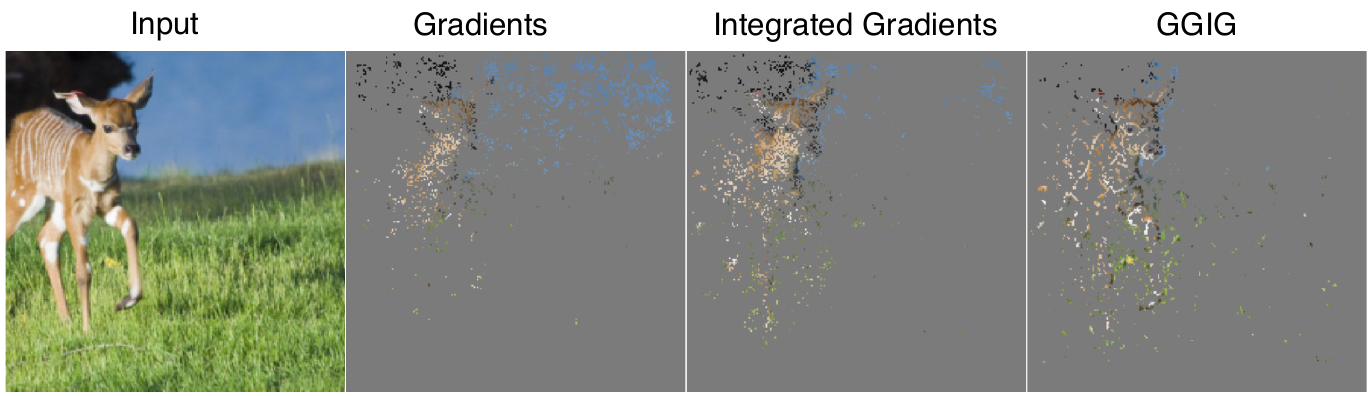}
  \caption{Comparison of gradients, integrated gradients and GGIG based on top 5\% pixel attributions for the "gazelle" prediction. In GGIG explanation, the details of the animal are clearly noticeable and discarded the undesirable regions from the explanation.  }
   \label{fig:example}
\end{figure}

These post-hoc analysis methods, specifically the ones that leverage model gradients to explain each sample, are known as saliency techniques. Vanilla gradients (GRAD)~\citep{baehrens2010explain, simonyan2013deep} and IG, two of the most popular methods, can be noisy ~\citep{montavon2017explaining, samek2016evaluating, smilkov2017smoothgrad, sturmfels2020visualizing}. We show an example of this noise in \figureautorefname~\ref{fig:example}. GRAD specifically violates both \emph{implementation invariance} and \emph{sensitivity}. IG also has specific noise relating to the averaging or integrating over the interpolation path. Other sources include the large curvature of the network's decision function \citep{dombrowski2019explanations}, numerical approximation of integration \citep{kapishnikov2021guided}, and baseline choices \citep{xu2020attribution, sturmfels2020visualizing}. In order to both conform to the two aforementioned properties and reduce noise, we suggest that the gradients can be improved by leveraging more aspects of the model space than the exact sample loss or a linear interpolation of the sample as is found in IG. Specifically, geometric properties of the loss landscape, which have previously been utilized to strengthen model performance and robustness ~\citep{garipov2018loss, gotmare2018using, izmailov2018averaging, entezari2021role}.

\begin{figure}[!htbp]
    \includegraphics[width=1.0\linewidth]{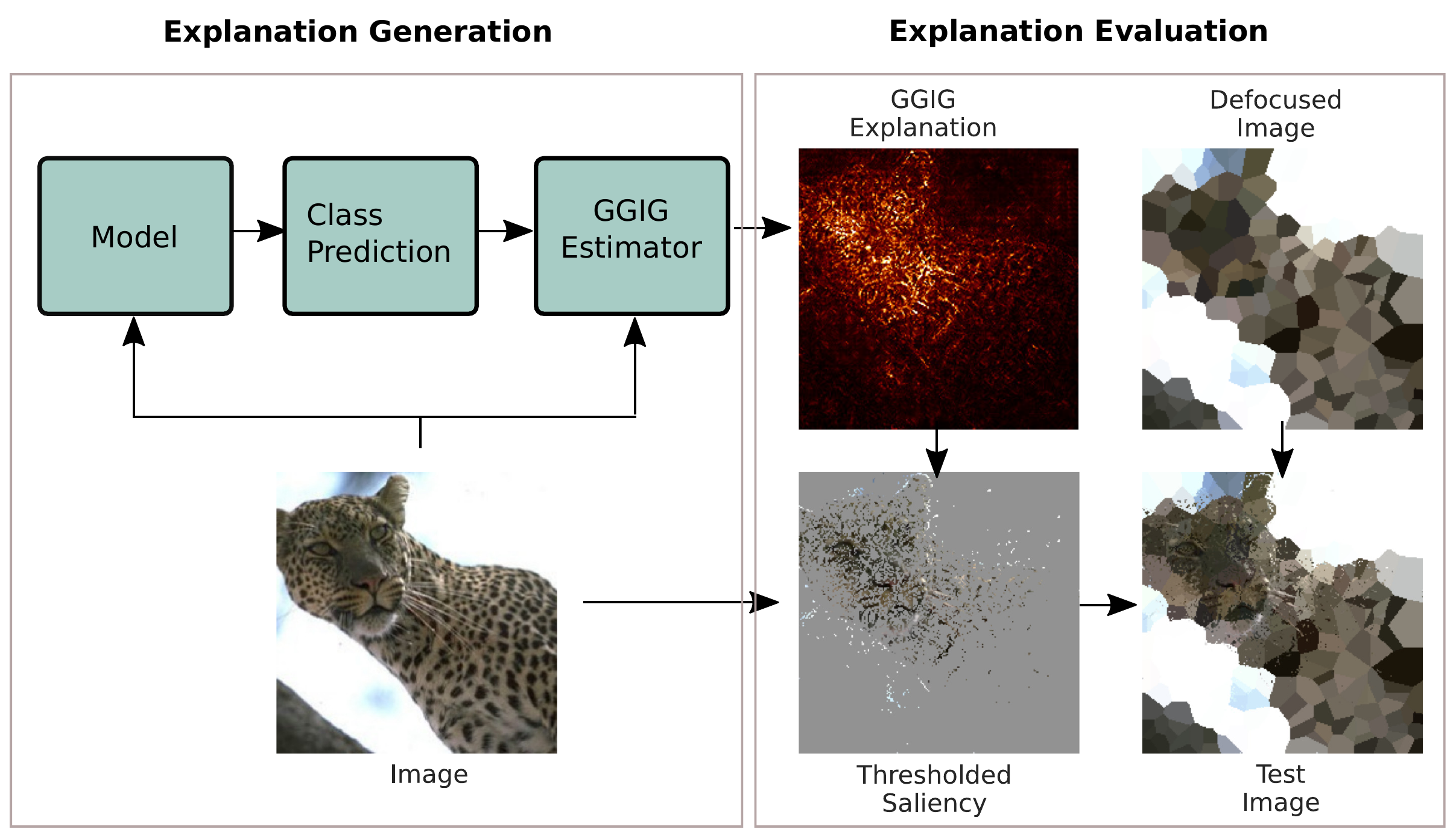}
  \caption{Overall pipeline of the work. First, we generate explanations for each prediction. Secondly, for quantitative evaluation, we start with a complete defocused image and combine with only salient pixels (thresholded at $x\%$) to create a test image. Finally, we feed the model with the test image and measure both accuracy and relative softmax scores.}
   \label{fig:eval_pipeline}
\end{figure}

In this paper, we propose \emph{Geometrically Guided Integrated Gradients (GGIG)}, an algorithm that builds on top of traditional IG to reduce noise. We expand IG by ascending through the loss space by maximizing the class-specific logit, much the way class activation maximization~\citep{couteaux2019towards} works. This ascension, or maximization of the class logit, allows GGIG to find the gradients that are most discriminative, which we suggest is a valuable property for any interpretability mechanism. After describing the method, we empirically show that GGIG improves the quality and robustness of the saliency maps. \figureautorefname~\ref{fig:eval_pipeline} describes the general pipeline of the work.

Our main contributions are as follows:

\begin{itemize}
\item We propose an interpretability method, called \emph{Geometrically Guided Integrated Gradients}, that starts with linear path as used in IG and finds the path that enhances the class activation for the underlying prediction.

\item We also propose a model perturbation sanity check, also called {\bf $\sigma$-perturbation} that we think all explanation methods should satisfy. 

\item We show that the proposed method offers better saliency maps for different datasets and architectures when assessed through visual inspection and quantitative metrics. 
\end{itemize}

\section{Related Work}
\smallskip
\noindent

The obfuscatory nature of DL models is well documented and has been a popular research problem for over a decade. Many studies have proposed solutions with varying quality, costs, and benefits ~\citep{baehrens2010explain, simonyan2013deep,
    shrikumar2017learning, smilkov2017smoothgrad,
    sundararajan2017axiomatic, xu2020attribution, selvaraju2017grad,
    springenberg2014striving, bach2015pixel, montavon2017explaining,
    hooker2019benchmark, adebayo2018local, zeiler2014visualizing}. Gradient-based methods, also referred to as \emph{visualization methods}~\citep{ras2022explainable}, are easy to implement and applicable to all models that use gradient descent. 

GradCAM \citep{selvaraju2017grad} identifies the focal regions pretty well, and maps are highly predictive. However, the saliency maps are blobby \citep{kapishnikov2019xrai} and usually expand around the actual objects. While IG has its own specific problems, several recent studies have refined IG attributions because this method has many desirable properties. \citet{kapishnikov2019xrai} proposed a region-based attribution method, called XRAI, that mainly refines the IG attribution based on attribution density. However, XRAI requires a way to cluster the input features, which may not be available for different data modalities.  \citet{kapishnikov2021guided} proposed another method, called GIG, which provides an adaptive path method based on input, baseline, and the model. This method starts at the baseline and selects only those pixels with the lowest partial derivatives to take closer to the next interpolation point, thus avoiding the gradient accumulation from saturation regions. In other words, it constitutes the path based on dynamic projections of the linear interpolation path. 

To reduce the inherent noise in the saliency maps,  we may utilize some useful loss landscape properties as observed in several studies~\citep{garipov2018loss, gotmare2018using, izmailov2018averaging, entezari2021role} to design reliable interpretability methods. 

\section{Geometrically Guided Integrated Gradients (GGIG)}
\label{ggig}

\begin{wrapfigure}{R}{0.5\textwidth}
\centering
\includegraphics[width=0.36\textwidth]{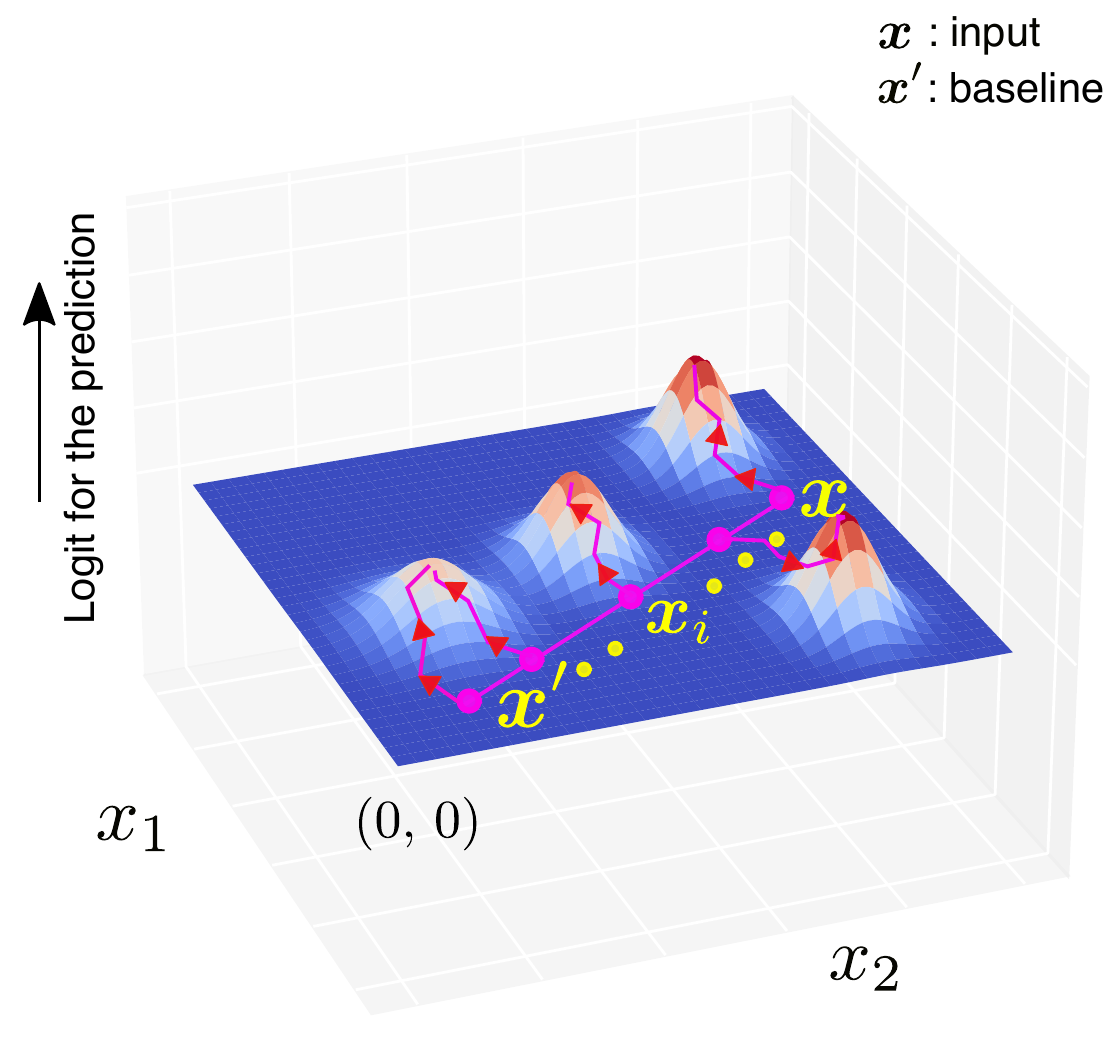}
\caption{\label{fig:main_idea}The basic idea of ``Geometrically Guided Integrated Gradients." It takes a baseline $\vx'$, creates a linearly spaced path to the actual input $\vx$. From each interpolation point, it follows a trajectory in the loss landscape that maximizes class activation.}
\end{wrapfigure}

We propose a method called GGIG, which incorporates the idea of \emph{path methods}~\citep{sundararajan2017axiomatic} and enhances the quality of the attribution by analyzing the local loss behavior. \figureautorefname~\ref{fig:main_idea} shows the schematic diagram illustrating the functional mechanism of GGIG.

\begin{algorithm}
\caption{Geometrically Guided Integrated Gradients (GGIG)}\label{alg:ggig}
\begin{algorithmic}[1]
\Procedure{\texttt{COMPUTE\_GGIG\_EXPLANATION}}{$F=\text{model logit function}$, $\vx=\text{sample}$, \newline\phantom{\textbf{procedure} \texttt{COMPUTE\_GGIG\_EXPLANATION}(}$n=\text{interpolation points}$, $\vx'=\text{baseline}$, \newline\phantom{\textbf{Procedure} \texttt{COMPUTE\_GGIG\_EXPLANATION}(}$m=\text{ascending iteration}$, $lr=\text{learning rate}$}

\State gradients $G \gets \{\}$

\For{$k \gets 0$ to $n$}
    \State $\vx_{k}^{0} \gets \vx' + \frac{k}{n}(\vx - \vx')$
    \For{$j \gets 0$ to $m-1$}
        \State $\vx_{k}^{j+1} = \vx_{k}^{j} + lr \times \nabla F(x_{k}^{j})$
        \State $G \overset{+}{\leftarrow} \nabla F(x_{k}^{j})$
    \EndFor
\EndFor
\State $\ve = \max(G)$  \Comment{pixel-wise maximum attribution from all gradient maps}
\State $\textbf{return} \, \ve$\Comment{$\ve$ is GGIG explanation} 
\EndProcedure
\end{algorithmic}
\end{algorithm}

Like IG, GGIG starts from the baseline $\vx'$ and constitutes a linearly interpolated path. However, instead of accumulating gradients along the path, it updates the path gradients in the direction where the model maximizes class activations. GGIG thus maximizes class activation for each of the linearly interpolated points. The procedural steps for GGIG are shown in {\bf Algorithm} \ref{alg:ggig}. We hypothesize that the prediction curve in the vicinity of $\vx$ holds important information about the interaction between model $\mathsf{f}$ and input $\vx$.

Let $F: \sR^{d} \rightarrow \sR$ be defined as the mapping from the input space to the class-specific logit and $\vb_0, \dots, \vb_n$ be a set of n linearly interpolated points for each sample between the baseline, $b_0$ and the exact sample $\vx$, where $\vb_i = \vb_0 + \frac{i}{n}(\vx - \vb)$. For each $\vb_i$, we compute a form of gradient ascent for $m$ iterations over the given interpolated sample (as opposed to the model), defined as: $b_{i}^{j+1} = b_{i}^{j} + lr \times \nabla F(b_{i}^{j})$ where $j$ is the incremental ascent over the sample and $lr$ is the learning rate. This ascending mechanism allows the model to take the direction where logit values are enhanced. All gradients over the ascension, $\nabla F(b_{i}^{j})$, are collected, and the max $\nabla F(b_{i}^{j})[p]$ (for each pixel $p$), where $i = 0, 1, 2, \dots, n$ and $j = 0, 1, 2, \dots, m$, is selected as the final attribution for the pixel $p$.





IG \citep{sundararajan2017axiomatic} has many desirbale properties and we have used IG as the baseline to compare the proposed method. We used slightly different formulation of IG. Generally, IG uses interpolation technique to integrate importance at different discrete intervals between uninformative baseline, say $\bar{\vx}$ and the input $\vx$, to give an integrated estimate of feature importance. IG based feature importance is computed as:
\begin{equation}
\ve = (\vx - \bar{\vx}) \times \sum_{i = 1}^{k} \frac{\partial \mathcal{F}(\bar{\vx}+\frac{i}{k} \times (\vx - \bar{\vx}))} {\partial \vx} \times \frac{1}{k}
\end{equation}
The ultimate estimate $\ve$ depends on the value of $k$ (number of intervals) and the choice of a suitable uninformative baseline $\bar{\vx}$. The traditional integrated gradients scale raw attributions (operand on the right of the multiplication operator) by $\vx- \vx'$ (operand on the left). 

{\bf Element-wise multiplication is misleading:}
 \citet{adebayo2018sanity} observed that element-wise multiplication could be misleading. This misleading happens mainly because the input dominates the product even with drastic changes in gradient vectors. So, the interpretability methods with this element-wise multiplication component in their formulation can provide input-dominant explanations that may deceive human understanding. 

 Furthermore, \citet{ancona2019gradient} suggested that this point-wise multiplication was initially justified to sharpen the gradient explanations; however, it is better justified when a measure of salience is a priority over mere sensitivity. In this paper, we were more interested in the sensitivity of features rather than their marginal salience to the target score. For all of these valid reasons, we did not multiply the integrated gradient with $\vx - \vx'$ to avoid input dominance from the explanations. Moreover, we did not consider other gradient-based methods like $\text{grad} \odot \text{input}$ for the same reason. 

The baseline can be problematic for correct attributions \citep{kapishnikov2019xrai, sturmfels2020visualizing}. \citet{kapishnikov2019xrai} addressed the issue of baseline by using both black $(0, 0, 0)$ and white $(1, 1, 1)$ baselines. While baseline may be an issue for traditional IG formulation, avoiding direct or modified (as in IG) element-wise multiplication resolves the baseline issue from this work. 

\section{Model Perturbation Sanity Check}

In this section, we propose a model perturbation sanity check for attribution methods. We refer to this perturbation as {\bf $\sigma$-perturbation}. This perturbation seeks two important properties:

\begin{myProp}
Let $M$ be a model with the parameter vector $\vw=[\evw_1, \evw_2, \dots, \evw_n]$ and $\sigma$ be the standard deviation of $\vw$. Let $\vx$ be the sample for which we are generating explanation. Let $S_{M}(\vx)$ be the probability score $M$ generates for the input $\vx$. Let $M'$ be a model obtained by perturbing the model $M$ using $\rvw' = \rvw + \mathcal{N}(\mathbf{0},\, \epsilon\mathbf{I})$, where $0 \le \epsilon \le \sigma$. For sufficiently large perturbation level $\epsilon \gg 0$, $S_{M'}(\vx)$ should reduce to $\nicefrac{1}{C}$, where $C$ is the number of classes.
\end{myProp}

\begin{myProp}
Let $\ve_{M}=E_{M}(\vx)$ be an explanation for the sample $\vx$ generated by the original model $M$ and $\ve_{M'} = E_{M'}(\vx)$ be an explanation for the same sample generated by the perturbed model $M'$. Also, let $S(\ve_1, \ve_2)$ be any similarity measure between two explanations $\ve_1$ and $\ve_2$. With gradual perturbation of the model $M$, the similarity between $\ve_M$ and $\ve_{M'}$ should evaporate accordingly. For sufficiently large perturbation level, i.e., $\epsilon \gg 0$, the $S(\ve_M, \ve_{M'}) \approx 0$.
\end{myProp}

We claim that all attribution methods should satisfy this sanity check to ensure their meaningfulness and sensitivity to the model parameters. We applied this sanity check to vanilla gradients, integrated gradients and GGIG. Interestingly, all of the methods satisfied the {\bf $\sigma$-perturbation} sanity check.

\begin{figure}[!htbp]
    \includegraphics[width=1.0\linewidth]{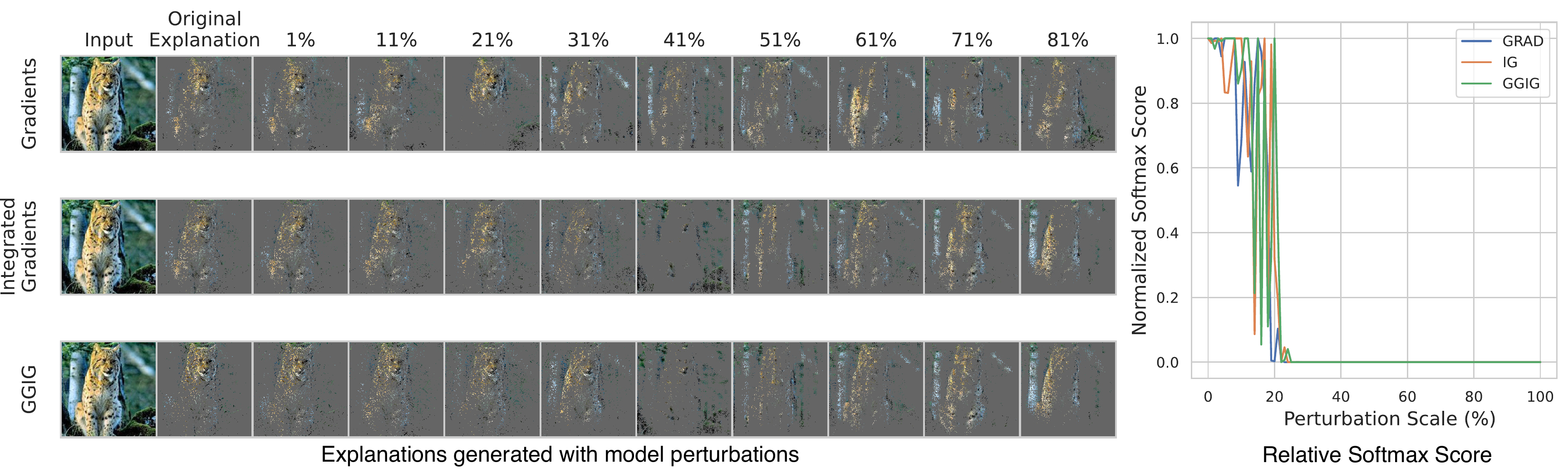} 
  \caption{{\bf Left:} Comparison of saliency-focused images thresholded at 10\% while we perturb the model. {\bf Right:} Relative {\bf softmax} score during model perturbation. As expected, the similarity with the original explanations breaks and softmax score drops with gradual increase in perturbation level.}
   \label{fig:model_perturbation_test}
\end{figure}

We report {\bf $\sigma$-perturbation} sanity check results in Figure \ref{fig:model_perturbation_test}. 

\section{Experiments}
{\bf Training on MNIST Dataset}
For MNIST~\citep{lecun1998mnist}, the model architecture was a CNN that consisted of two convolutional layers with (32, 64) filters of sizes (5, 5). Each convolutional layer is followed by a $2 \times 2$ max pooling layer and a ReLU activation. We fed the final convolution output to a fully connected network with 1024 input and 10 output units (softmax). 
We optimized the model using stochastic gradient descent (SGD) with a learning rate of 0.0004 and momentum of 0.9. The model was trained for 400 iterations with a mini-batch size of 64 and finally achieved an accuracy of 99.2\%. 

{\bf Post hoc explanation experiments}
For GGIG, we used a learning rate of 0.0001 for gradient ascent from each linear interpolation point between input $\vx$ and baseline $\vx' = \vzero$. For MNIST, we iterated the gradient ascent for 200 steps and noted the maximum sensitivity along the gradient ascent trajectory for each input. We display the saliency maps obtained on the MNIST dataset in Figure \ref{fig:mnist}.

{\bf Quantitative Evaluation on MNIST Dataset}
Visual inspection of explanation methods can be unreliable as it is possible to create adversarial samples ~\citep{goodfellow2014explaining, szegedy2013intriguing} that can fool the human eye, totally changing the model predictions. We perform different similarity measures between the maps and the input to understand the quality of the proposed methods on MNIST, namely, \emph{Spearman Rank Correlation, Weighted Jaccard Similarity, Structural Similarity, and Normalized (Reverse) Mean Square Error}. For quantitative evaluation, both data and maps were rescaled in the range [0, 1]. We assumed that the amount of information a method can capture about the structure and distribution of the input in the saliency maps directly determines its quality as an explanation method. We report the quantitative evaluation in \tableautorefname~\ref{sim_analyis_table}. Figure \ref{fig:mnist} in Appendix shows sample maps and the detailed results of quantitative evaluation. As we can observe, GGIG outperforms other methods. Comparatively, GRAD and IG retain little information about the numerical and structural association to the input. 

\begin{table}[!htbp]
\caption{Summary of Quantitative Evaluation on MNIST dataset}
  \label{sim_analyis_table}
  \centering
\begin{tabular}{ccccccccc}
\toprule
\multirow{2}[3]{*}{\makecell[c]{Saliency \\ Method}} & \multicolumn{2}{c}{\makecell[c]{Spearman Rank \\ Correlation}} & \multicolumn{2}{c}{\makecell[c]{Weighted Jaccard \\ Similarity}} & \multicolumn{2}{c}{\makecell[c]{Structural \\ Similarity}} & \multicolumn{2}{c}{\makecell[c]{Normalized \\ Mean Square Error}}\\
\cmidrule(lr){2-3} \cmidrule(lr){4-5} \cmidrule(lr){6-7} \cmidrule(lr){8-9} 
 & median & std & median & std & median & std & median & std \\
\midrule
GRAD & 0.525 &  0.048 & 0.210 & 0.035 & 0.167  & 0.040 & 0.413 & 0.107 \\
IG & 0.516 & 0.049 & 0.214 & 0.035 & 0.170 & 0.040 & 
0.414 & 0.108\\
GGIG & \textbf{0.533} & 0.056 & \textbf{0.332} & 0.041 & \textbf{0.244} &  0.043 & \textbf{0.540} & 0.099\\
\bottomrule
\end{tabular}
\end{table}

\begin{figure}[!htbp]
    \includegraphics[width=1.0\linewidth]{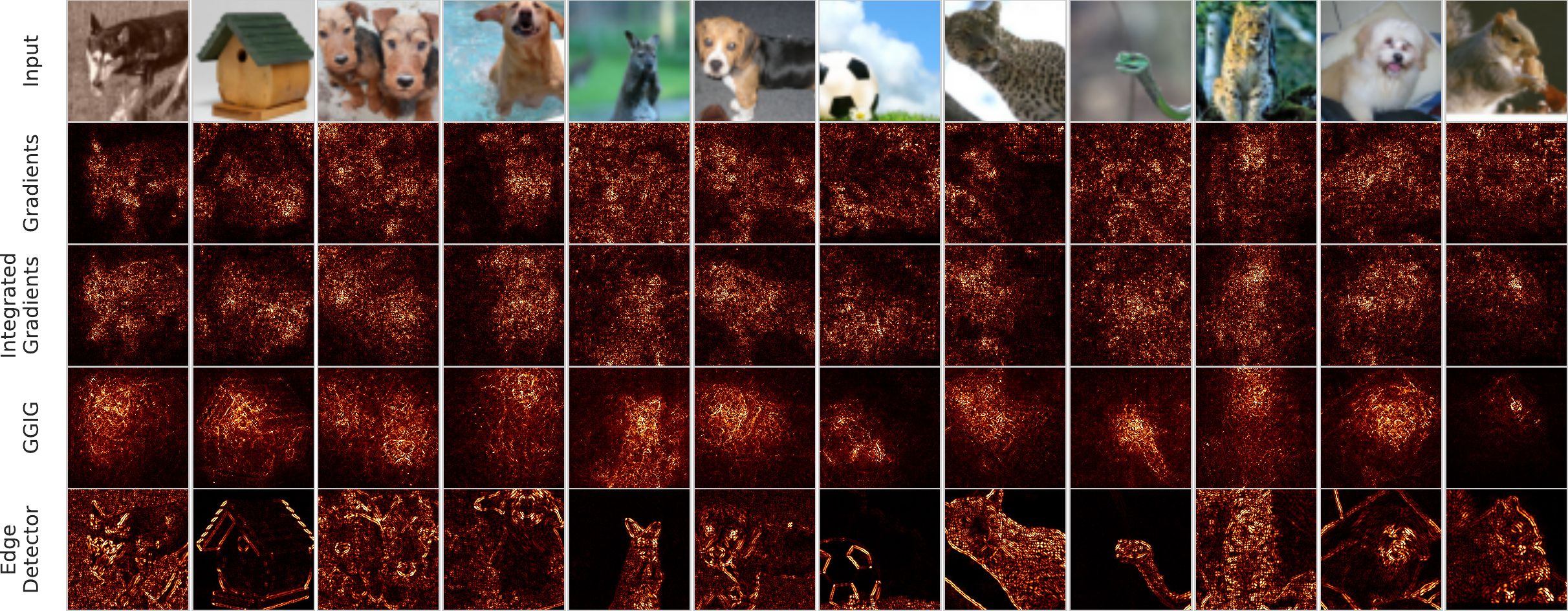}
  \caption{Sample saliency maps for ImageNet Validation set and ResNet101 model generated by gradients, integrated gradients, geometrically guided integrated gradients, and Edge Detector.}
   \label{fig:resnet101_valSet}
\end{figure}

{\bf Experiments on ImageNet:} We also evaluated the proposed methods using a subset of images from the ImageNet dataset~\citep{krizhevsky2012imagenet} and different pretrained models, namely ResNet-101~\citep{he2016deep}, and Inception V3~\citep{szegedy2016rethinking}. Though we found meaningful maps in every case, maps still vary in quality possibly due to their architectural differences. \figureautorefname~\ref{fig:resnet101_valSet} and \figureautorefname~\ref{fig:inception_testSet} show some maps (more results are in Appendix) produced using different saliency methods and pretrained models.

\begin{figure}[!htbp]
    \includegraphics[width=1.0\linewidth]{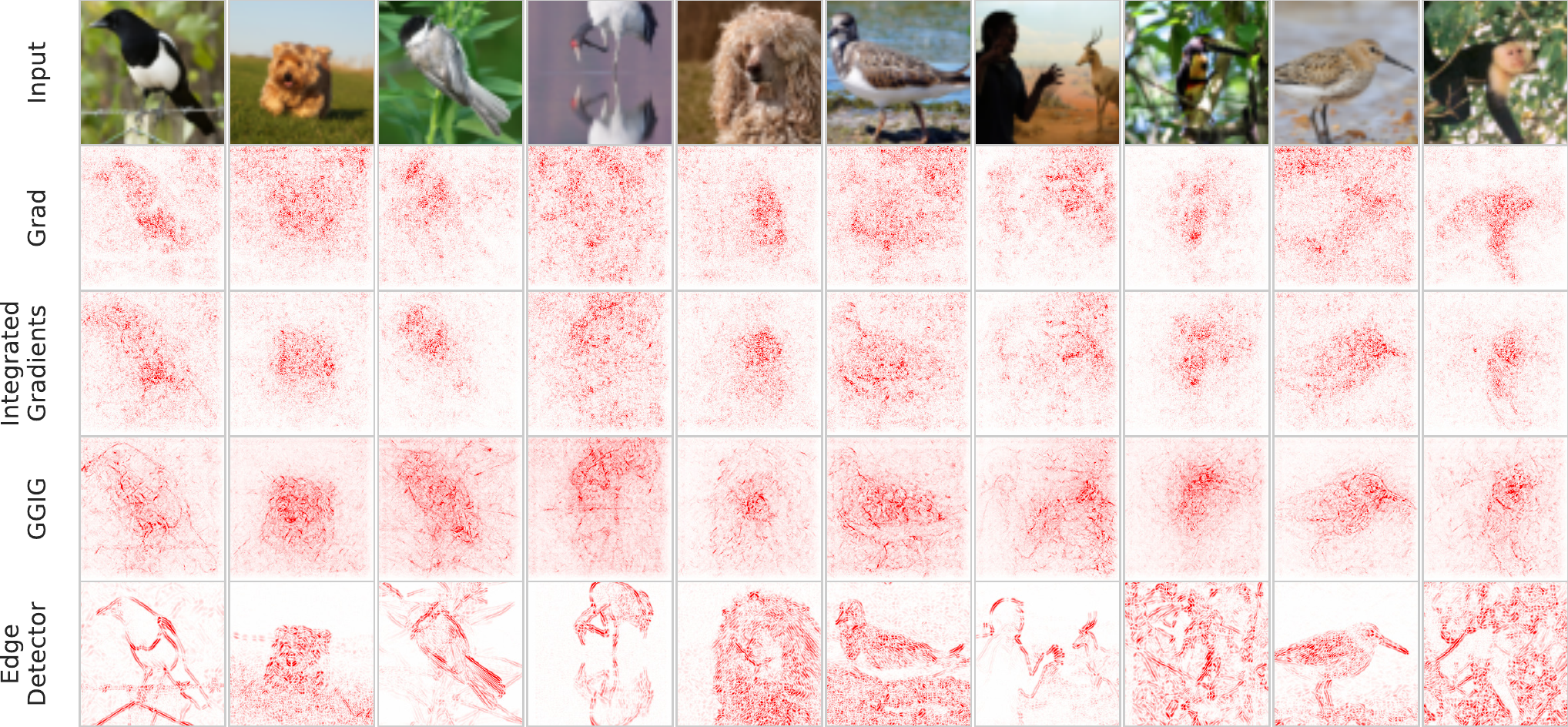}
  \caption{Comparative saliency maps for Inception V3 model and ImageNet dataset generated by GRAD, IG, GGIG, and Edge Detector. The maps obtained using GGIG are more discriminative and more clearly reveal the underlying structure of the class-associated objects.}
   \label{fig:inception_testSet}
\end{figure}

\begin{figure}[!htbp]
    \includegraphics[width=1.0\linewidth]{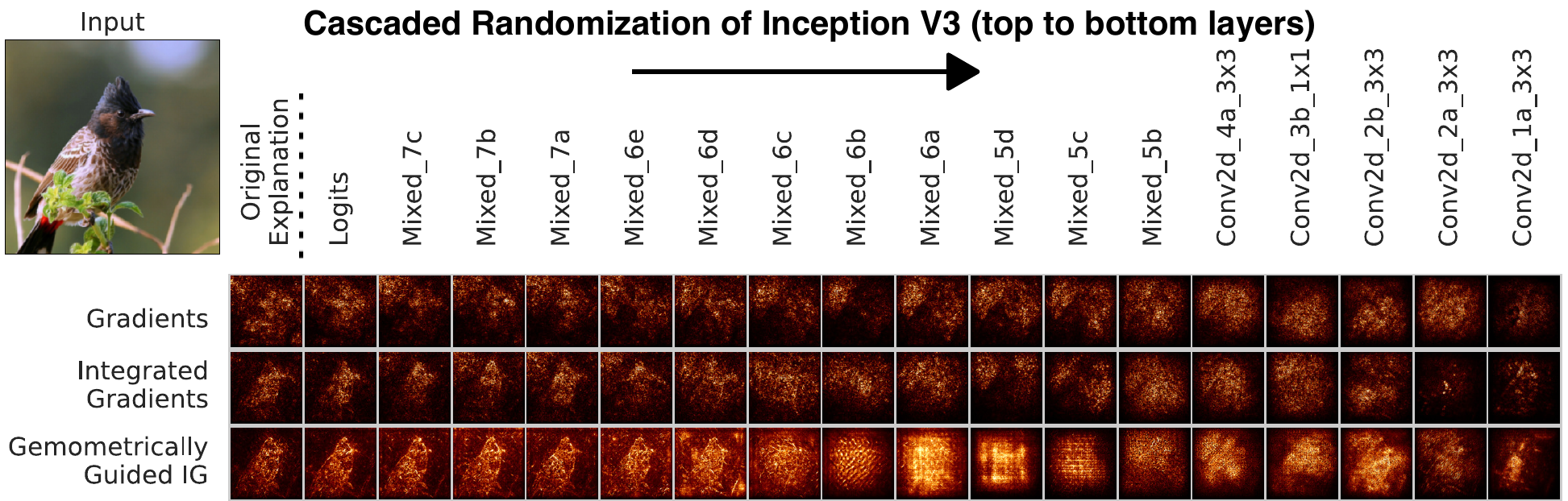}
  \caption{We show how different methods generate saliency maps when layers of the underlying model (Inception V3) are gradually randomized from top to bottom.}
   \label{fig:rand_test}
\end{figure}

{\bf Model Randomization Test:} We also performed a \emph{Model Randomization} test~\citep{adebayo2018sanity} (\figureautorefname~\ref{fig:rand_test}) to verify the sensitivity of the methods to the model parameters. To this end, we randomly reinitialized the weights and generated post-hoc explanations using the randomized model. The proposed method GGIG is as sensitive as GRAD and IG, suggesting that our method is highly sensitive to model parameters.

\begin{figure}[!htbp]
    \includegraphics[width=1.0\linewidth]{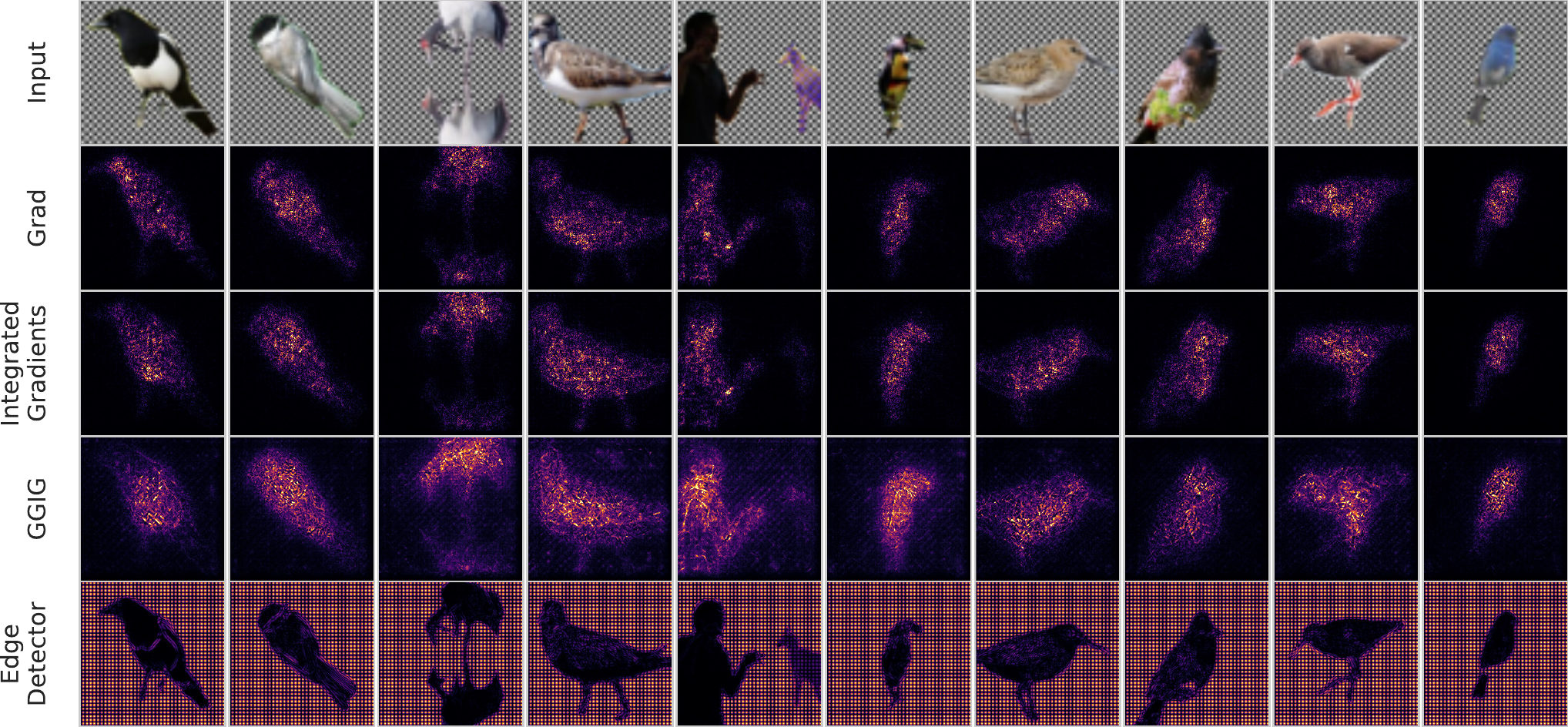}
  \caption{Background replacement experiments on ImageNet samples and Inception V3 model. While the traditional edge detector supposedly identifies all of the sharp changes and does not focus any specific attention to the actual objects, the saliency methods are still focusing on the actual objects for predictions.}
   \label{fig:bg_experiment}
\end{figure}

{\bf Saliency methods are not edge detectors:} \citet{adebayo2018sanity} observed that many saliency methods, including vanilla gradients and integrated gradients can appear like edge-detectors for \emph{1-Layer Sum-Pool Conv Model}. This assumption may only hold for shallow models. To analyze the edge-detector like behavior for deep models, we conducted an experiment where we replaced the original background of the images with a sharp-changing image. In particular, we assigned ImageNet samples a very different fixed background (a black and white checkerboard) and generated post-hoc explanations using GRAD, IG, and GGIG. We show the resulting explanations in \figureautorefname~\ref{fig:bg_experiment}. It is obvious from the resulting maps that the model used concepts, not merely edges from the training objects. Moreover, as expected, all the saliency methods ignored the background and only attributed the object's parts for prediction explanations. It is apparent that GGIG method retained learned concepts more accurately during post-hoc explanations. We think that for well-trained deep models, saliency methods no-longer function like an edge-detectors.  

\subsection{Evaluation of Attribution Quality}

\begin{figure}[!htbp]
    \includegraphics[width=1.0\linewidth]{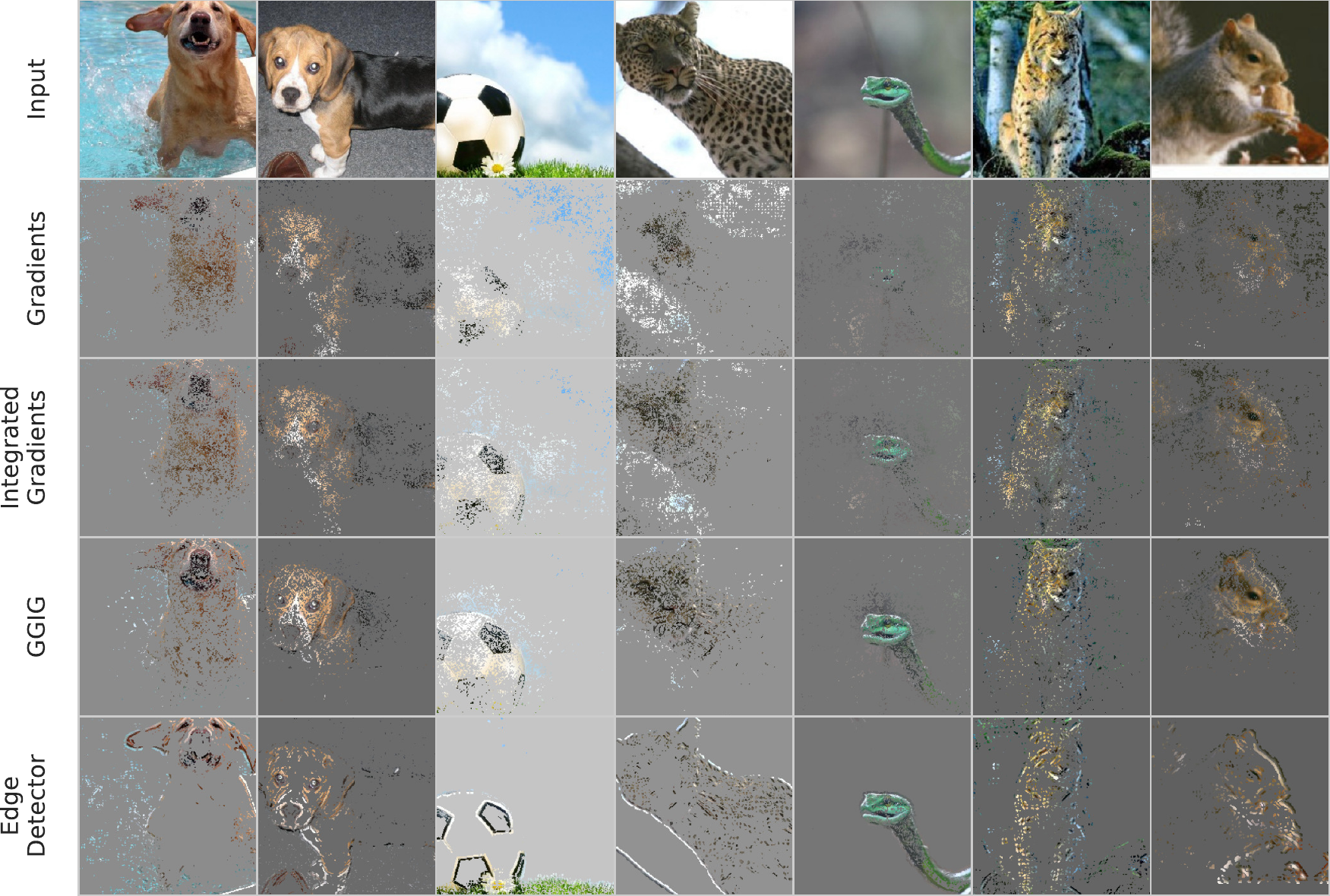}
  \caption{Comparison of saliency-focused images thresholded at 10\%. GGIG, compared to other methods, is more convincing from the human standpoint because it directly points to the discriminative parts of the image while ignoring redundant and unexpected regions.}
   \label{fig:comp_saliency_focused_images}
\end{figure}

{\bf Visual Analysis:} We show few sample explanations thresholded at 10\% in \figureautorefname~\ref{fig:comp_saliency_focused_images}. As we can observe, GGIG provides the least noisy explanations compared to other methods. As expected, the edge detector does not pay attention to discriminative regions. Rather, it captures only the sharp changes throughout the image. While GRAD and IG attribute lots of redundant or unexpected parts of the image, GGIG directly points to the most discriminative pixels of the image. Moreover, GGIG mostly avoids sharp changes in the image (see the explanation for the "leopard"), while GRAD and IG are highly susceptible to the edges. 

{\bf Quantitative Evaluation:} It is challenging to evaluate an interpretability method because of the lack of ground truth saliency or consensus metrics for proper evaluation. Several studies proposed different measures, such as \emph{Remove And Retrain (ROAR)} \citep{hooker2019benchmark}, \emph{Retain And Retrain (RAR)} \citep{rahman2022interpreting}, \emph{Accuracy Information Curves, Softmax Information Curves} \citep{kapishnikov2019xrai} to assess the quality of explanations.

The ROAR approach modifies the dataset by removing the features that received top attribution values from each sample. In practice, if the training data have sufficient (redundant) discriminative features \citep{sturmfels2020visualizing}, even after removing a significant number of features, the performance of the retrained model does not drop noticeably. In that scenario, ROAR fails to evaluate the feature relevance correctly. Retain and Retrain (RAR) method resolves the problem by retaining only the critical features instead of removing them. However, both ROAR and RAR methods are time-consuming as they require full retraining of the model. 

\citet{dabkowski2017real} proposed a metric, called Smallest Sufficient Regions (SSR), based on the notion of the smallest sufficient region capable of correct prediction. However, this metric is not suitable if the model is susceptible to the scale and aspect ratio of the object. Moreover, as this metric depends on rectangular cropping and reports results as a function of the cropped area, this approach highly penalizes if the saliency map is coherently sparse \citep{kapishnikov2019xrai}. Because, in that case, it may span a larger area of the image than the map, which is locally dense, even with the same number of pixels. SSR also imposes a severe challenge because masking creates a sharp boundary between the masked and salient region, causing an out-of-distribution problem for the model. 

\cite{kapishnikov2019xrai} proposed two metrics, called  \emph{Accuracy Information Curve} and \emph{Softmaxe Information Curve}, collectively called \emph{Performance Information Curve (PIC)}, to evaluate the saliency maps with the minimal out-of-distribution setting. We used SIC and AIC to evaluate our method and adopted a similar setup as used in \citep{kapishnikov2019xrai}. However, instead of using compressed image size as a proxy information level, we directly computed the entropy of the gray version of the saliency-focused image (test image). 

\begin{figure}[!htbp]
    \includegraphics[width=1.0\linewidth]{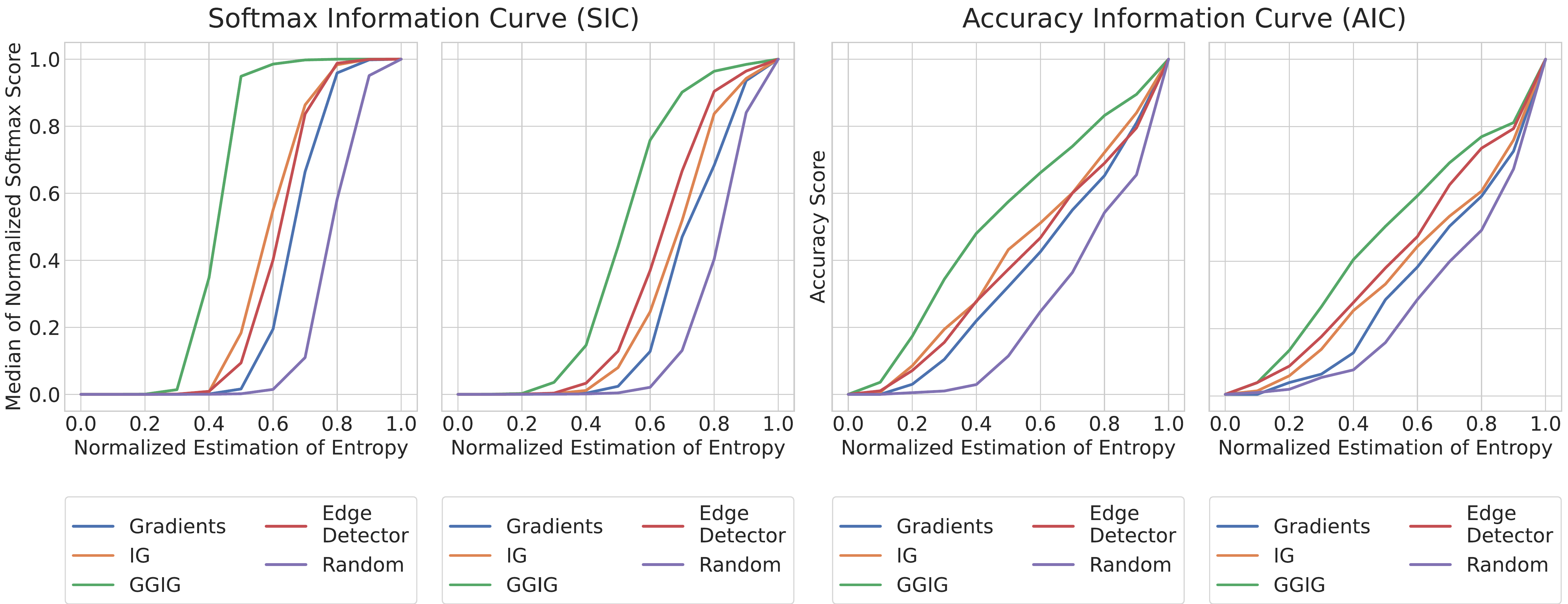} 
  \caption{{\bf Left:} Evalution of saliency methods using Softmax Information Curve (SIC) metric for Inception v3 \citep{szegedy2016rethinking} and ResNet101 \citep{he2016deep} on ImageNet test and validation images respectively {\bf Right:} Evaluation using Accuracy Information Curve (AIC) metric for the same models and datasets.  }
   \label{fig:quant_eval_imgnet}
\end{figure}

We show the quantitative evaluation results in \figureautorefname~\ref{fig:quant_eval_imgnet}. We started with a complete defocused image and gradually added salient pixels to form a saliency-focused image. We measure the entropy of the saliency-focused image relative to the complete blurred image and call it a normalized entropy, which forms the $x$-axis. We feed saliency-focused images to the original model and report median softmax scores and accuracy calculated over all images, which forms the $y$-axis. We show the evaluation procedure in \figureautorefname~\ref{fig:eval_pipeline}.

\begin{table}[!htbp]
\caption{Summary of SIC and AUC Evaluation}
  \label{sic_auc_table}
  \centering
\begin{tabular}{lcccc}
\toprule
\multirow{2}[3]{*}{Method} & \multicolumn{2}{c}{Inception v3} & \multicolumn{2}{c}{ResNet 101} \\
\cmidrule(lr){2-3} \cmidrule(lr){4-5}
 & SIC & AIC & SIC & AIC \\
\midrule
Gradients & 0.337 & 0.364 & 0.271 & 0.32 \\
Integrated Gradients & 0.409 & 0.415 & 0.315 & 0.361 \\
GGIG & \textbf{0.576} & \textbf{0.523} & \textbf{0.469} & \textbf{0.472} \\
Edge Detector & 0.384 & 0.394 & 0.355 & 0.404 \\
Random & 0.201 & 0.241 & 0.189 & 0.258 \\
\bottomrule
\end{tabular}
\end{table}

\section{Conclusion}
In this paper, we propose an interpretability method called GGIG that leverages the model's dynamic behavior in parameter space to enhance the quality of the feature attributions. GGIG starts with a linearly interpolated path as IG and uses gradient ascent to explore how the model interacts with input in the neighborhood. We believe that by looking into the parameter space from different interpolation points, GGIG captures important information about model vs. input interaction. We show the effectiveness of the proposed method through visual inspection and several quantitative metrics across some popular architectures and datasets. As demonstrated, we expect that this work provides insights toward building a more robust interpretability method through model and input interaction. 

\begin{ack}
This work was supported by the NIH grants RF1MH121885, R01MH123610, R01EB006841 and NSF grant 2112455.
\end{ack}


\bibliographystyle{plainnat}
\bibliography{interpretability}



\newpage
\appendix

\section{Quantitative Evaluation of GGIG on MNIST Dataset}
\begin{figure}[!ht]
    \includegraphics[width=\textwidth]{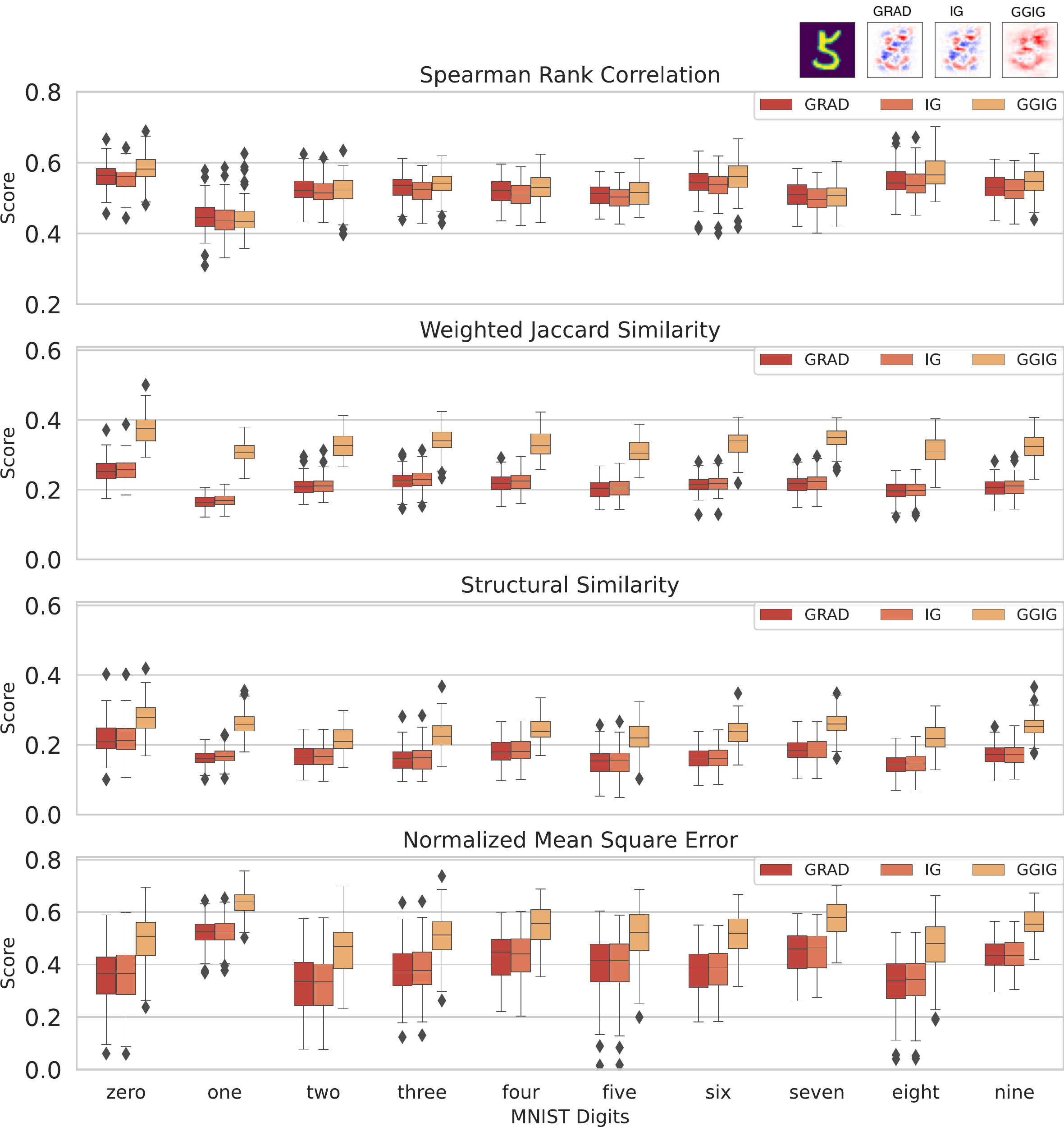}
  \caption{{\bf Top: } Selected maps for MNIST samples generated using different methods. {\bf Bottom:} Quantitative evaluations using different correlation and similarity metrics. It is obvious that the maps produced by GGIG have higher structural and numerical correlations with the input. Precisely, GGIG, as an explanation method, performs the best by a large quantitative margin.}
  \label{fig:mnist}
\end{figure}

\newpage
\section{Saliency Maps (ImageNet Dataset)}
\subsection{Architecture: ResNet-101}
\label{app:D2}
\begin{figure}[!htbp]
    \includegraphics[width=1.0\linewidth]{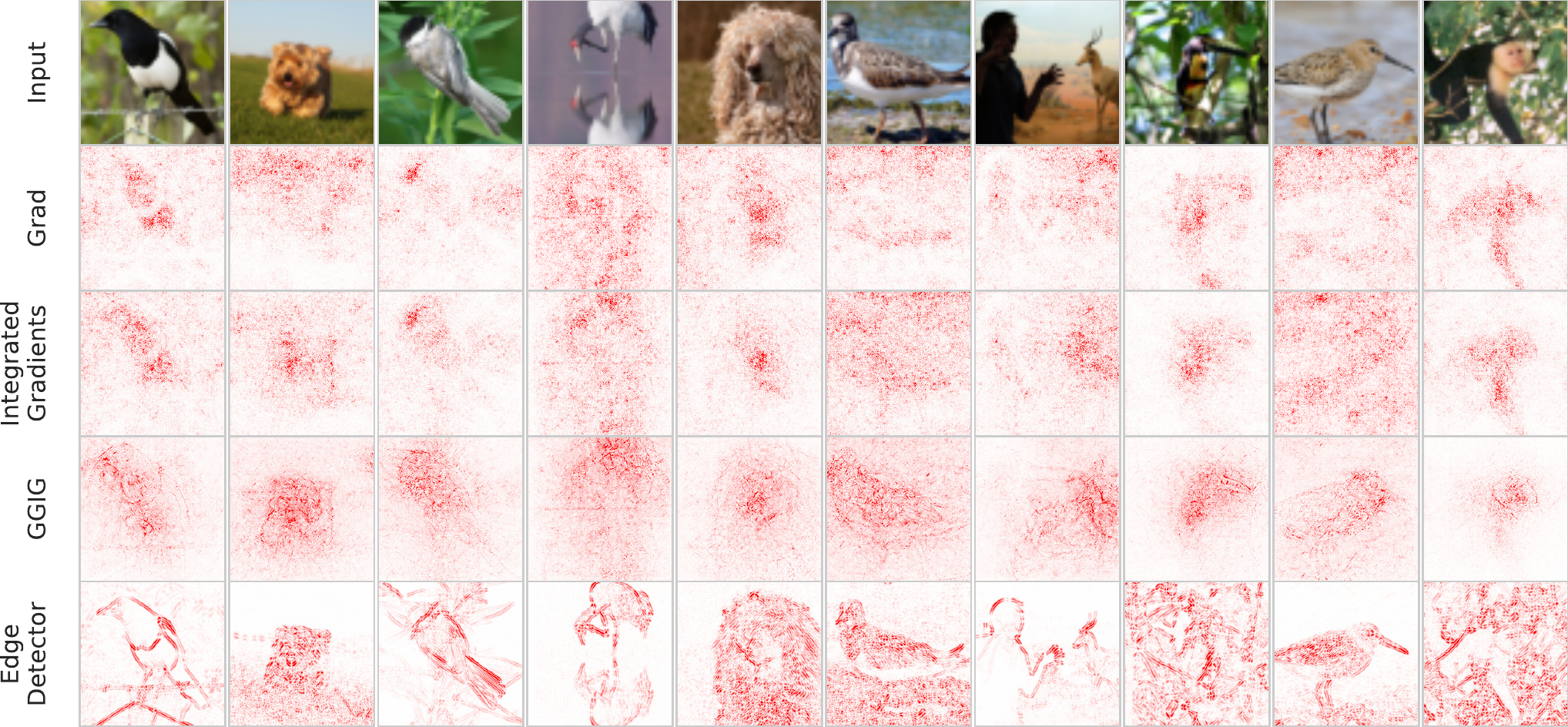}
    \label{fig:imgnet_resenet101_bwr}
    \vspace{-0.5cm}
  \caption{Explanations (saliency maps) of ResNet-101 predictions for ImageNet samples generated by GRAD, IG, and GGIG methods. As we can see, the GGIG maps have strong correlation and similarity with the corresponding input images.}
\end{figure}

\vspace{2cm}
\begin{figure}[!htbp]
    \includegraphics[width=1.0\linewidth]{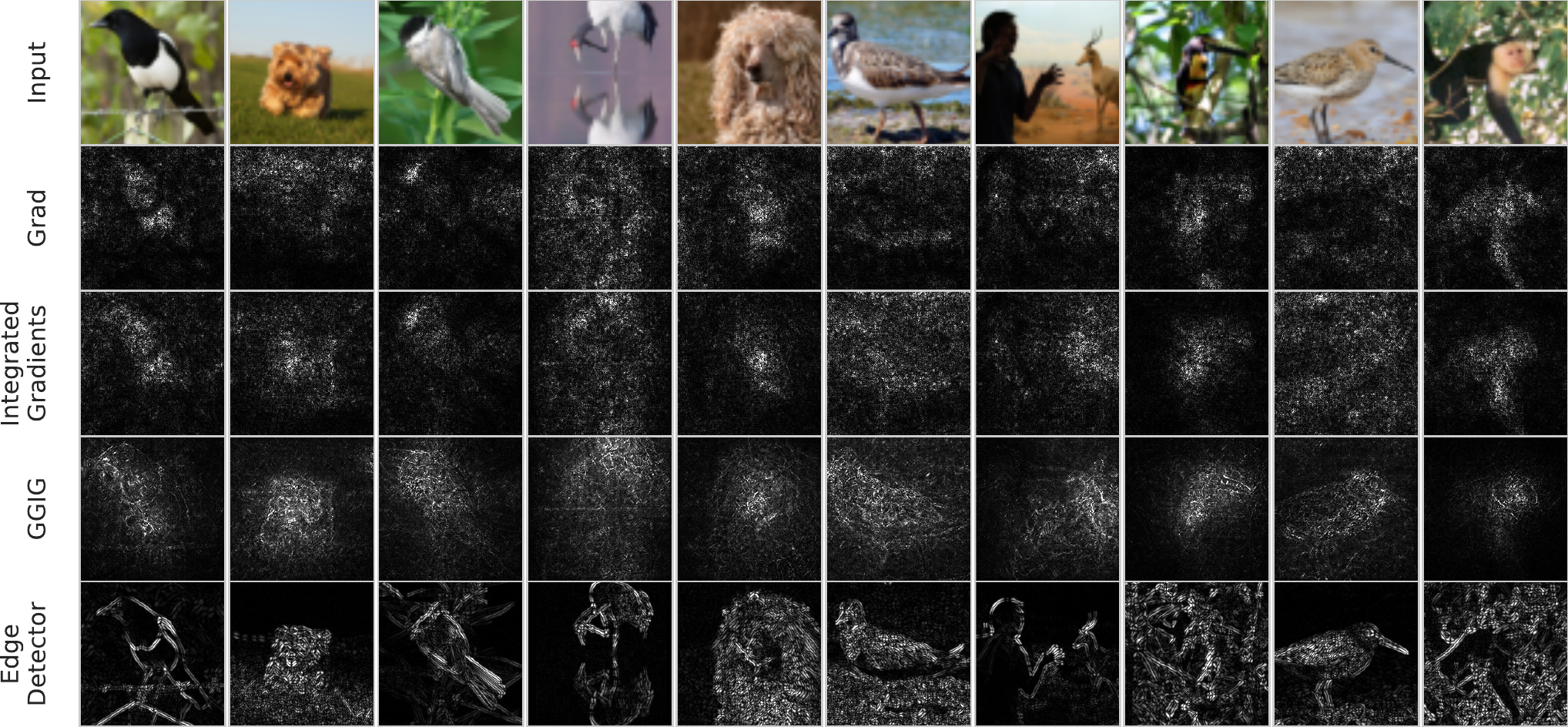}
    \label{fig:imgnet_resenet101_gray}
    \vspace{-0.5cm}
  \caption{Visualization of saliency maps obtained from Resnet-101 using 'gray' color scheme.}
\end{figure}

\newpage
\subsection{Architecture: Inception V3}
\label{app:D3}

\begin{figure}[!htbp]
    \includegraphics[width=1.0\linewidth]{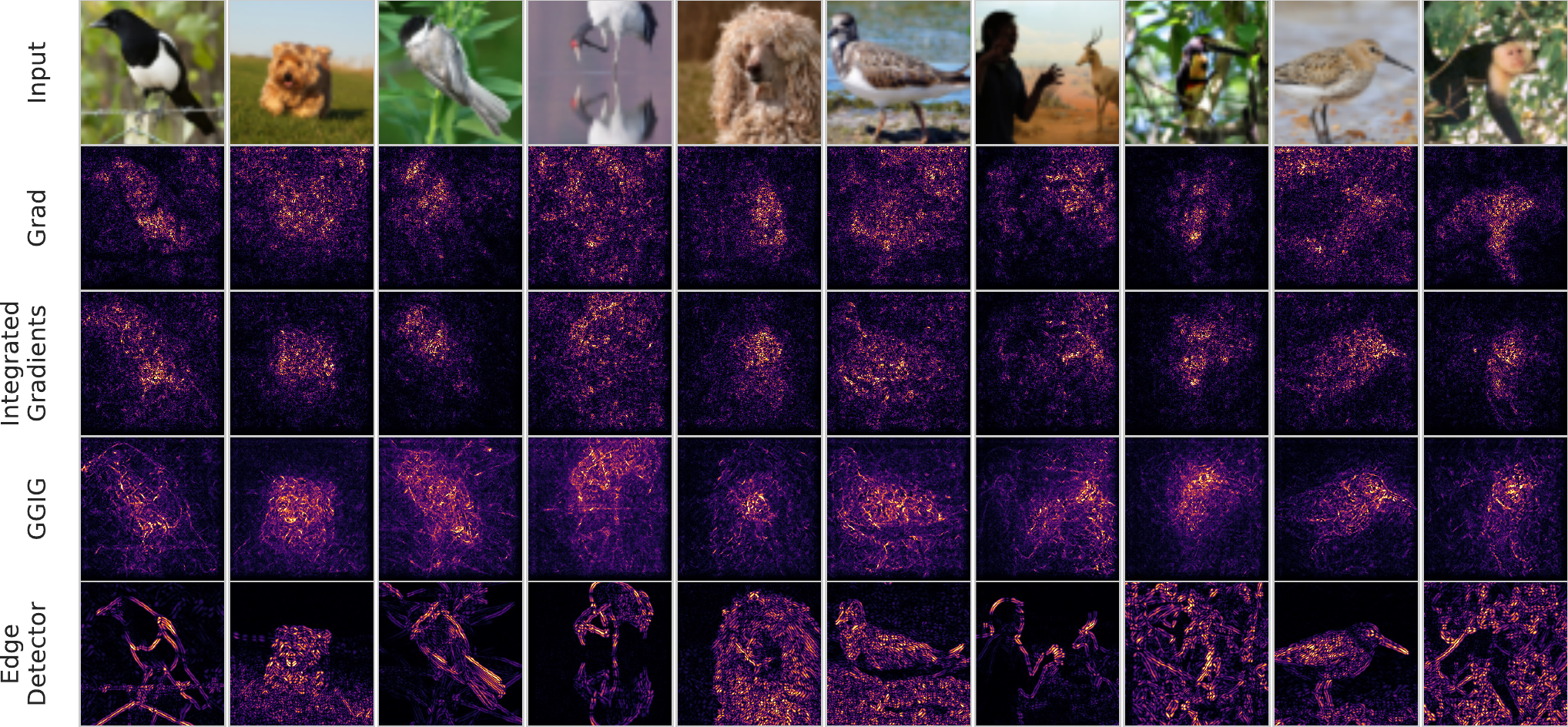}
    \label{fig:imgnet_inception_inferno}
    \vspace{-0.5cm}
  \caption{Visualization of saliency maps using 'inferno' color scheme for selected ImageNet samples. It is evident that the GGIG method consistently captures the useful concepts as used by the model for predictions.}
\end{figure}
\vspace{2cm}
\begin{figure}[!htbp]
    \includegraphics[width=1.0\linewidth]{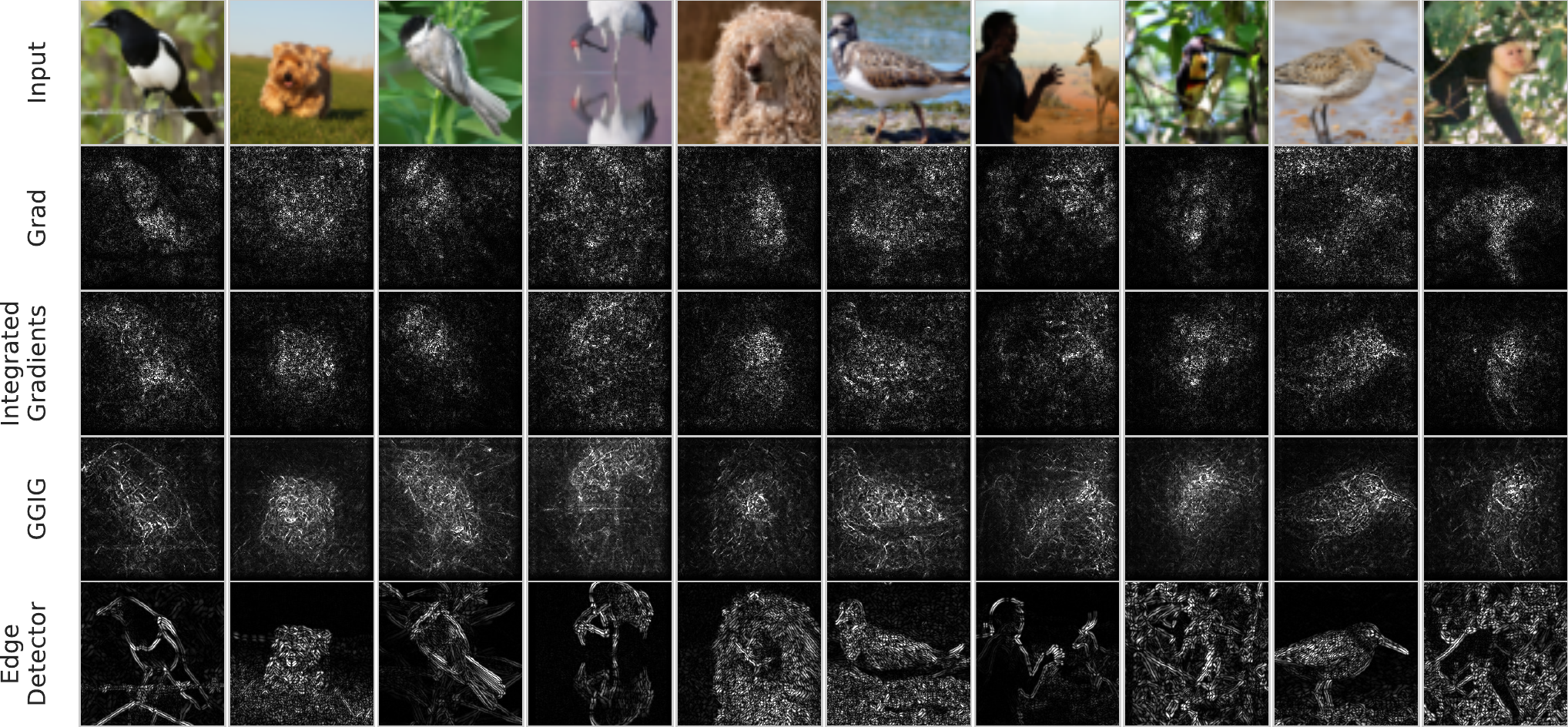}
    \label{fig:imgnet_inception_gray}
    \vspace{-0.5cm}
  \caption{Visualization of saliency maps using 'gray' color scheme for selected ImageNet samples.}
\end{figure}

\newpage
\begin{figure}[!htbp]
    \includegraphics[width=1.0\linewidth]{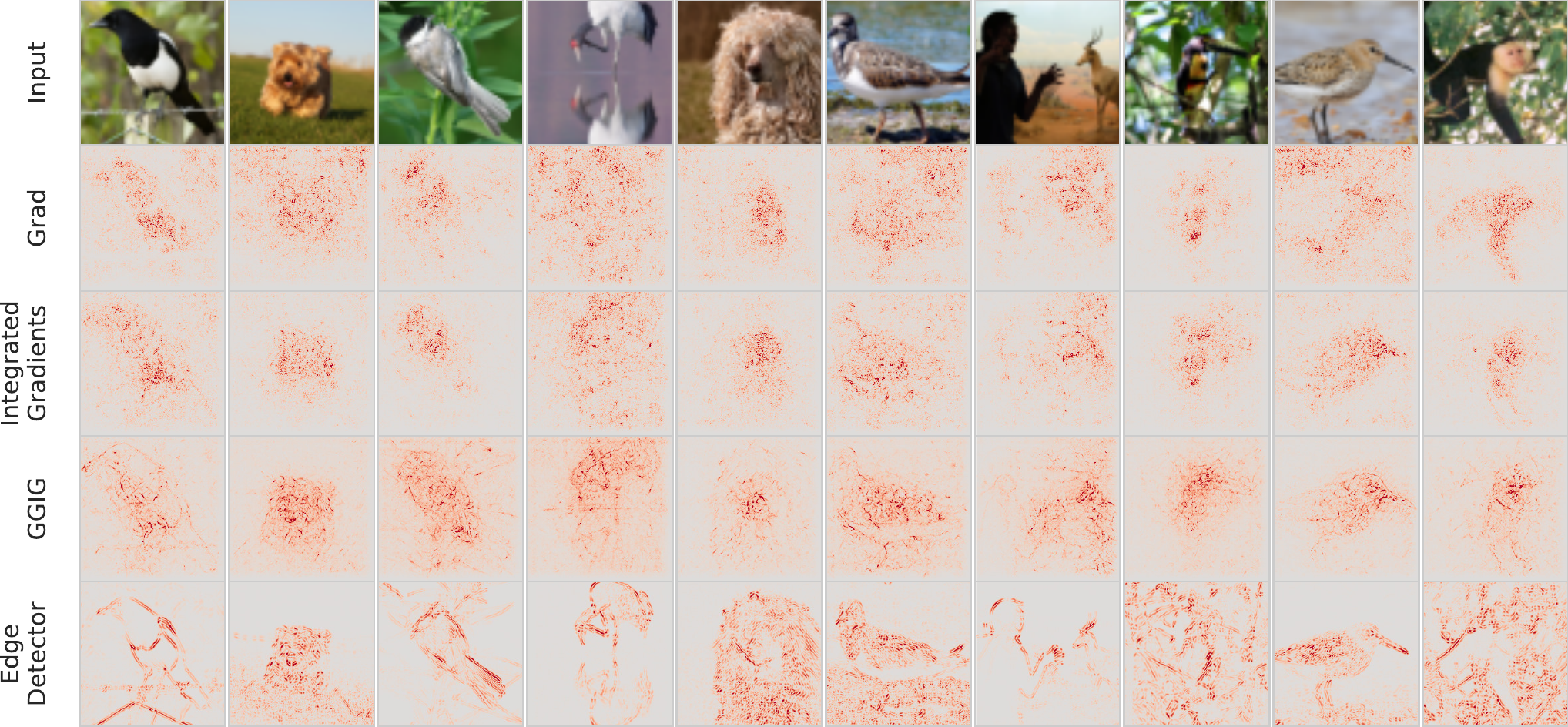}
    \label{fig:imgnet_inception_coolwarm}
    \vspace{-0.5cm}
  \caption{Visualization of saliency maps for selected ImageNet samples using 'coolwarm' color scheme.}
\end{figure}
\vspace{2cm}
\begin{figure}[!htbp]
    \includegraphics[width=1.0\linewidth]{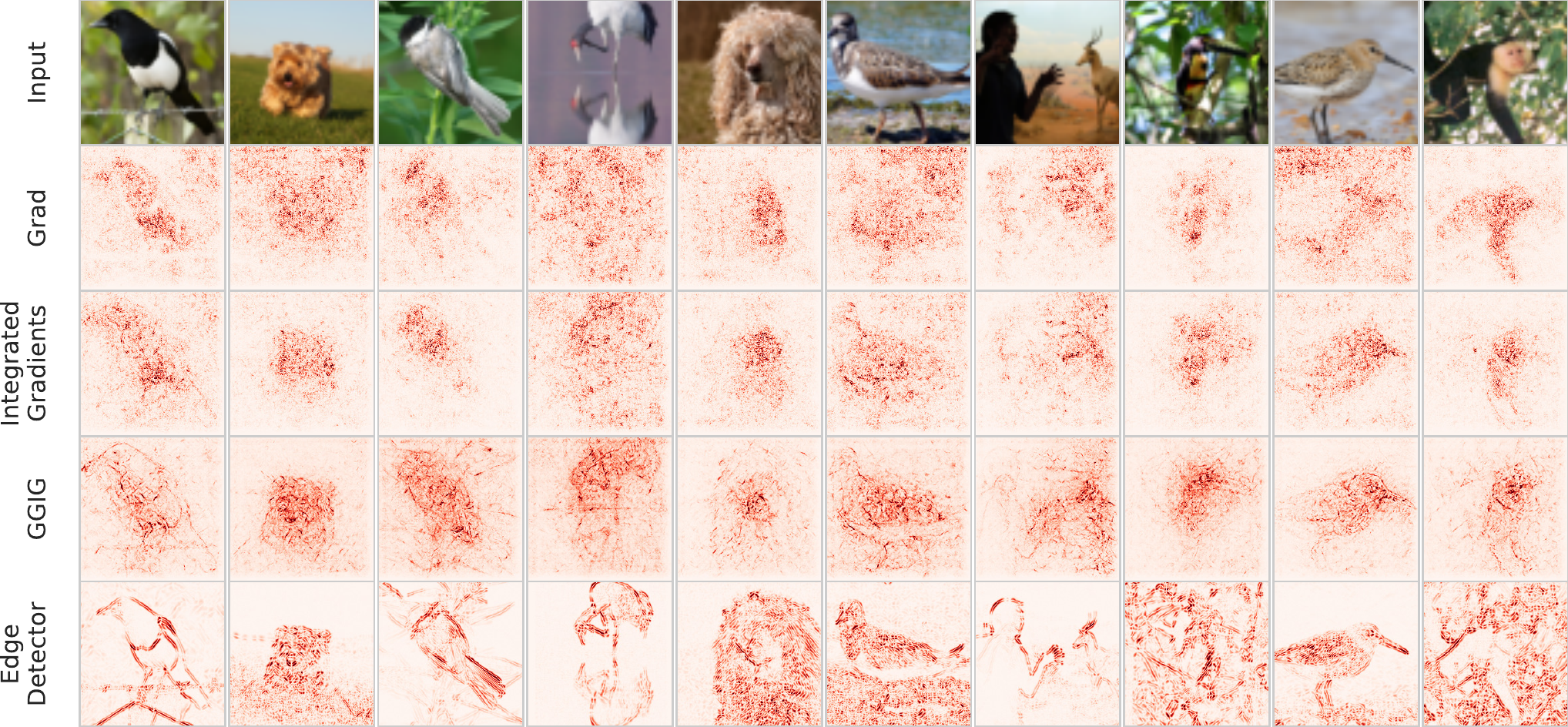}
    \label{fig:imgnet_inception_reds}
    \vspace{-0.5cm}
  \caption{Visualization of saliency maps for selected ImageNet samples using 'Reds' color scheme.}
\end{figure}

\newpage

\begin{figure}[!htbp]
    \includegraphics[width=1.0\linewidth]{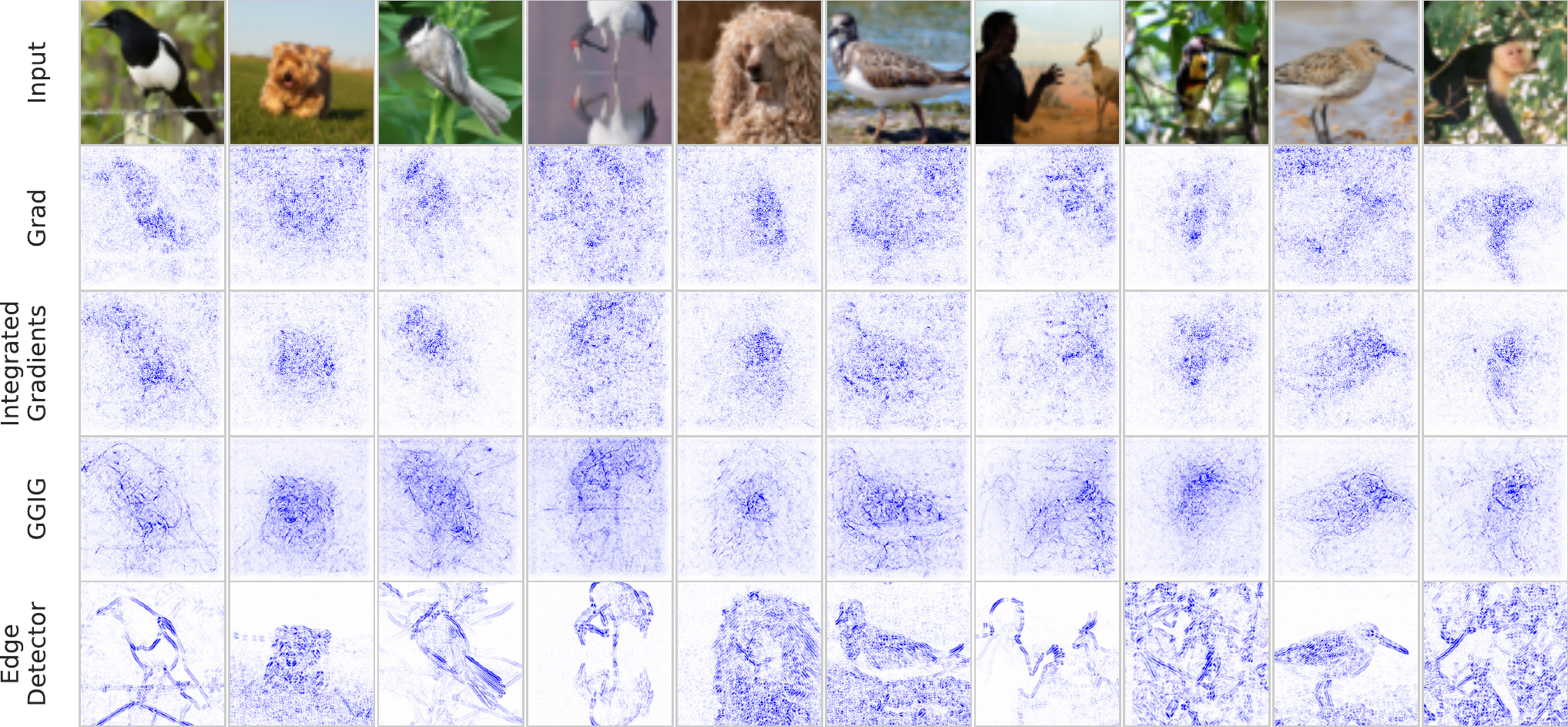}
    \label{fig:imgnet_inception_bwr_r}
    \vspace{-0.5cm}
  \caption{Visualization of saliency maps for selected ImageNet samples using 'reverse bwr' color scheme.}
\end{figure}

\subsection{Background Replacement Experiments }
\label{app:D4}
\begin{figure}[!htbp]
    \includegraphics[width=1.0\linewidth]{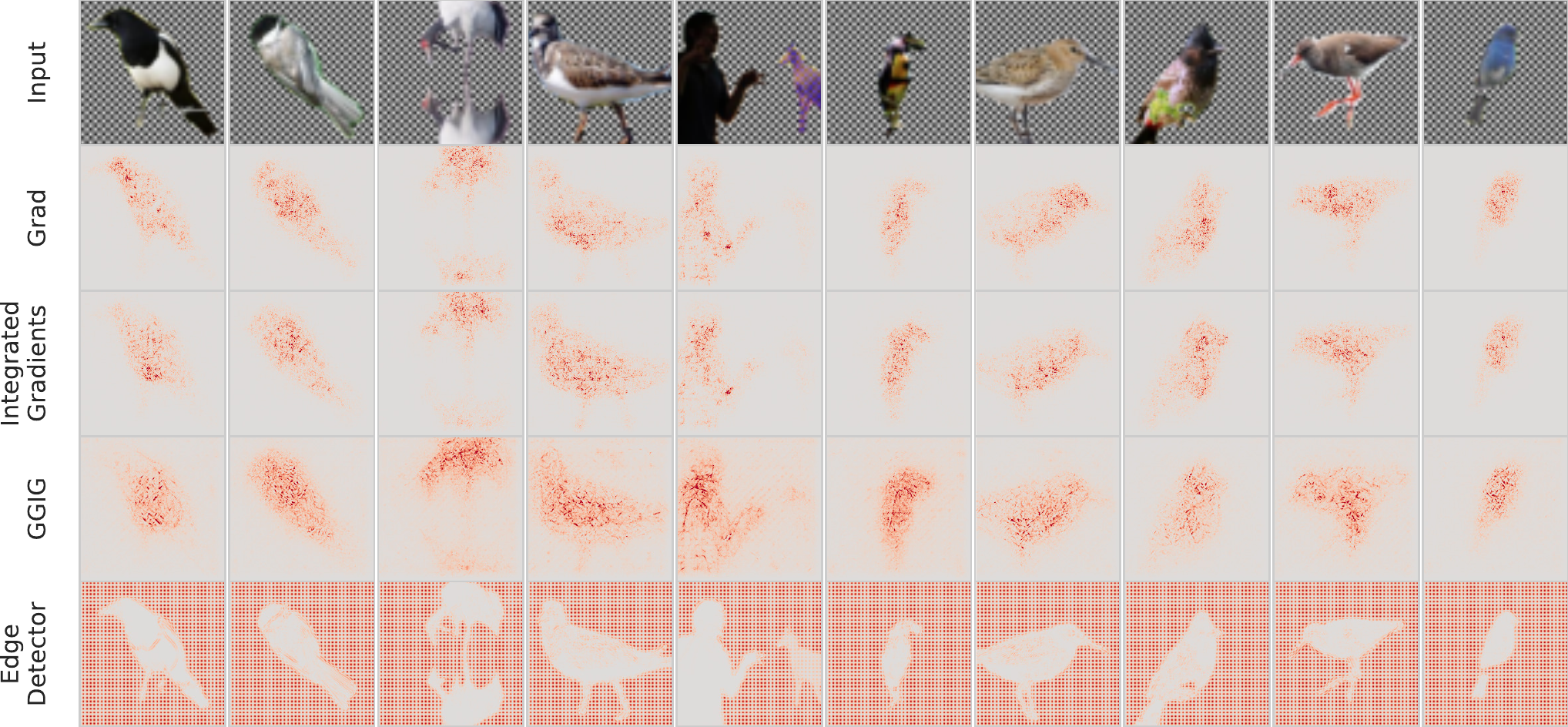}
    \label{fig:imgnet_bg_replacement}
  \caption{Visualization of background replacement experiments using 'coolwarm' color scheme.}
\end{figure}

\begin{figure}[!htbp]
    \includegraphics[width=1.0\linewidth]{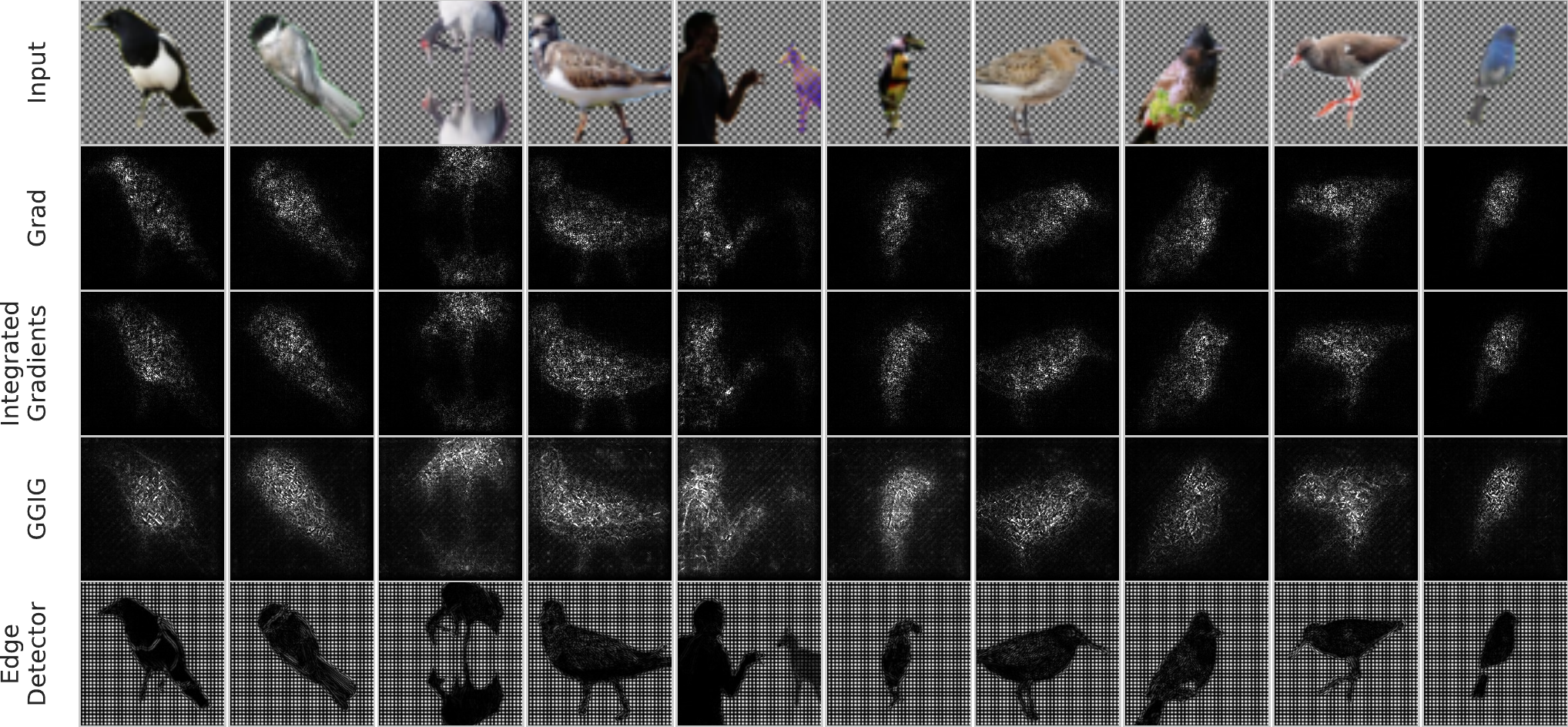}
    \label{fig:imgnet_bg_replacement_2}
  \caption{Visualization of background replacement experiments using 'gray' color scheme.}
\end{figure}

\newpage 
\section{Cascaded Randomization}

\label{app:D}

\begin{figure}[!htbp]
    \includegraphics[width=1.0\linewidth]{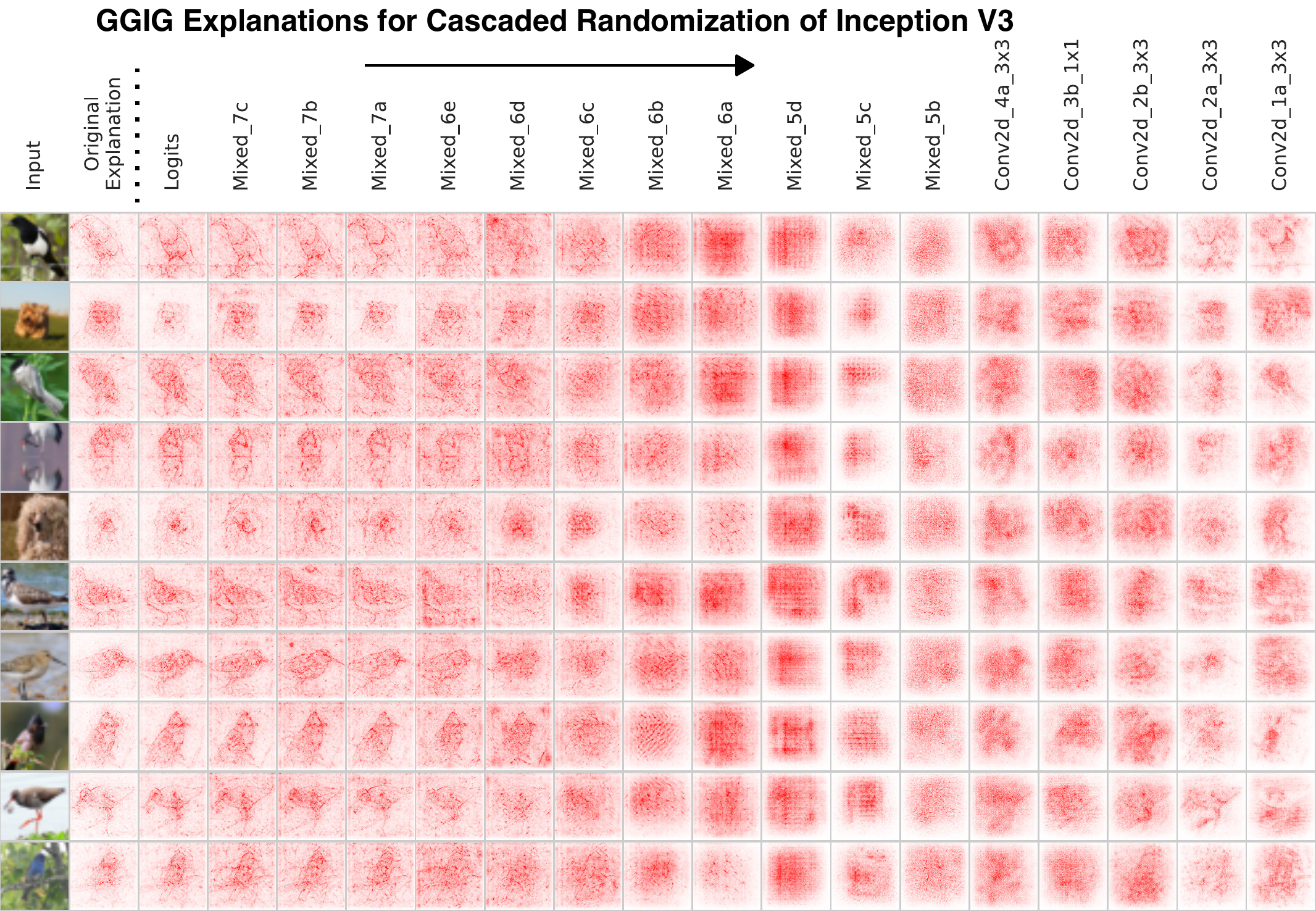}
    \label{fig:rand_test_multiple}
    \vspace{-0.5cm}
  \caption{'bwr' color scheme visualization of GGIG explanations of several samples for \emph{Cascaded Randomization} of Inception V3.}
\end{figure}

\begin{figure}[!htbp]
    \includegraphics[width=1.0\linewidth]{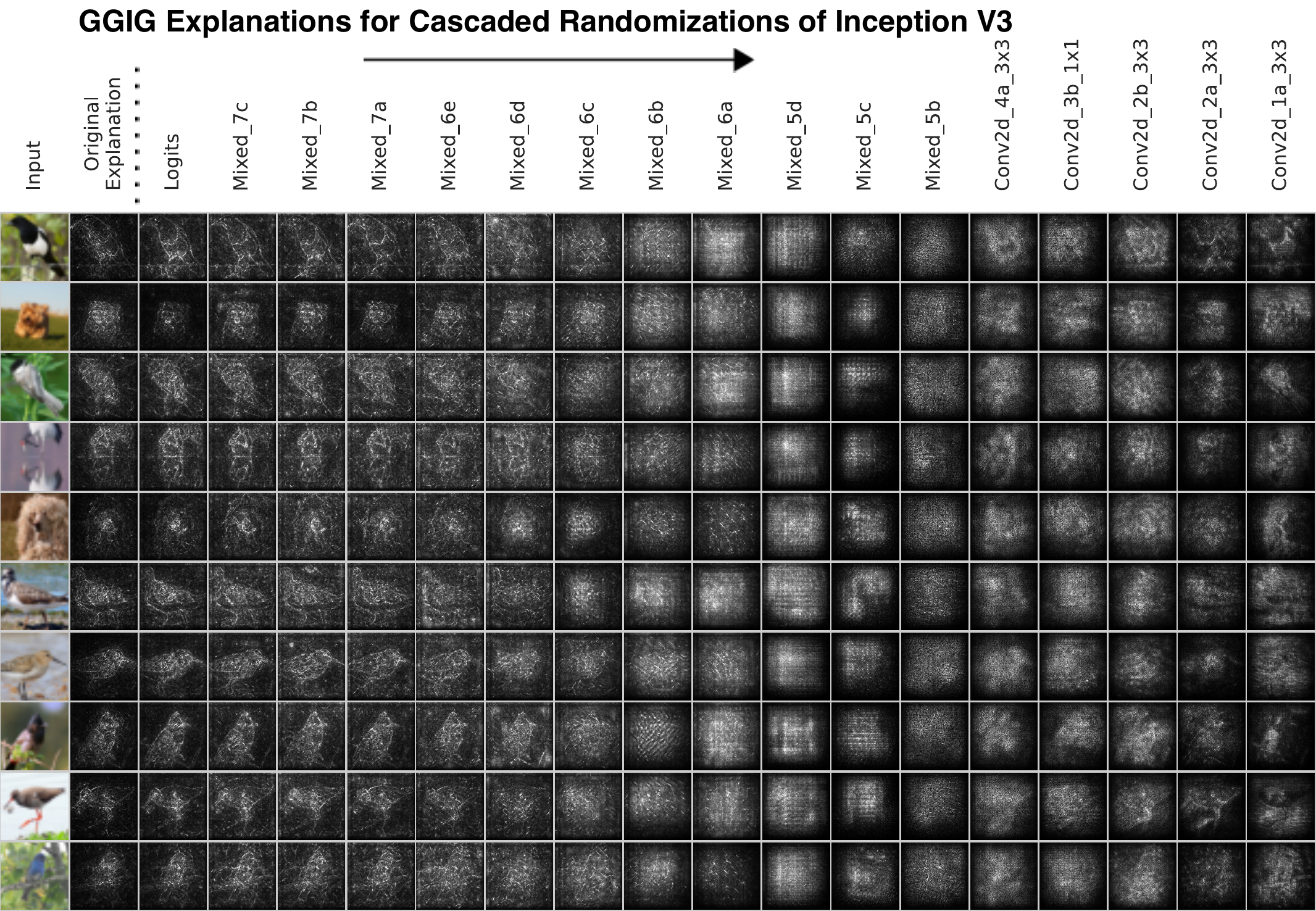}
    \label{fig:rand_test_multiple_2}
    \vspace{-0.5cm}
  \caption{'gray' color scheme visualization for GGIG explanations of several samples for \emph{Cascaded Randomization} of Inception V3.}
\end{figure}

\begin{figure}[!htbp]
    \includegraphics[width=1.0\linewidth]{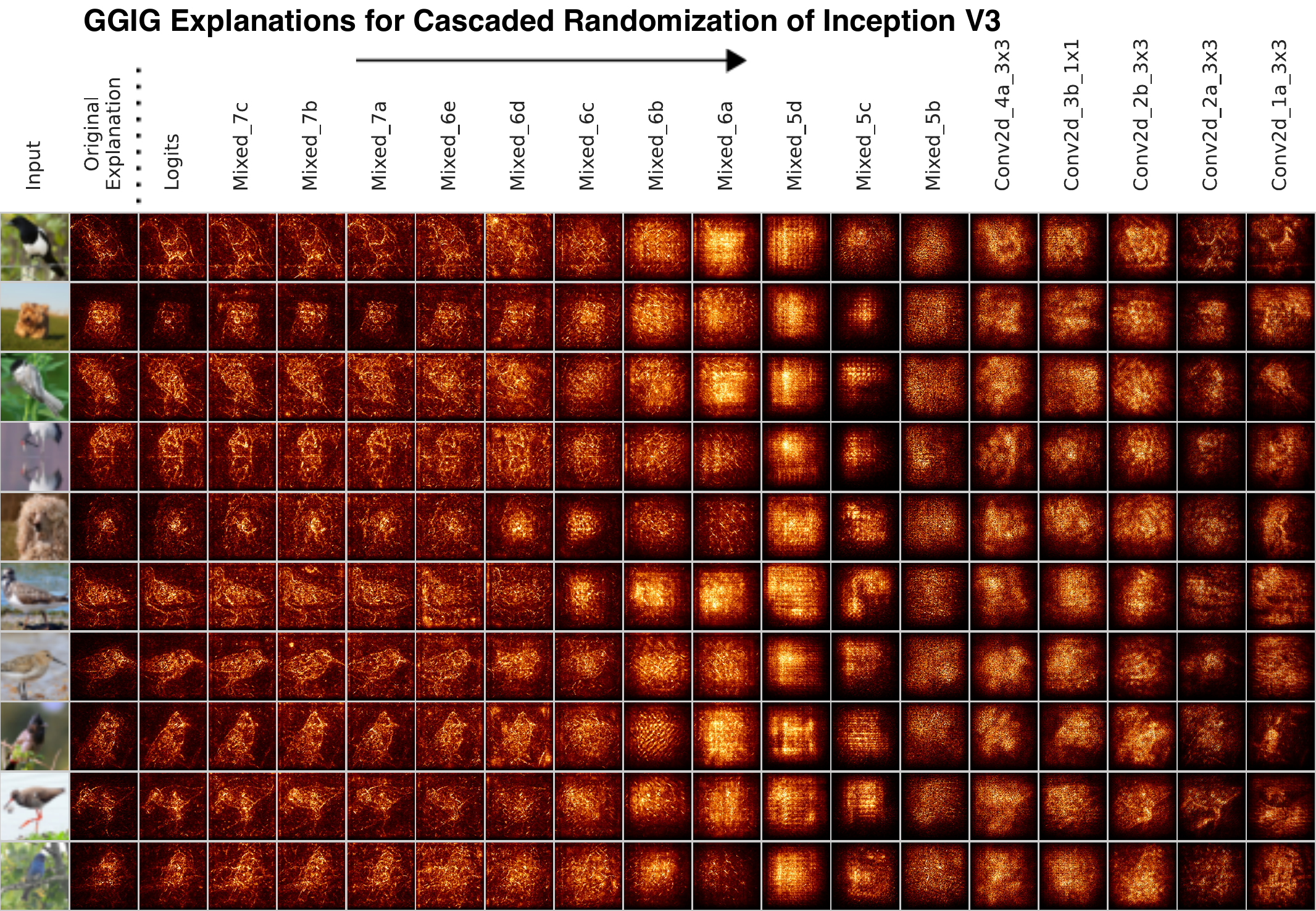}
    \label{fig:rand_test_multiple_3}
    \vspace{-0.5cm}
  \caption{'afmhot' color scheme visualization for GGIG explanations of several samples for \emph{Cascaded Randomization} of Inception V3.}
\end{figure}

\end{document}